%% file: main.tex
\documentclass[11pt]{article}

\usepackage[]{emnlp2021}

\pdfoutput=1
\usepackage{times}
\usepackage{latexsym}
\usepackage{enumitem}
\usepackage{booktabs}
\usepackage[title]{appendix}

\usepackage{caption}
\usepackage{subcaption}

\usepackage[utf8]{inputenc}

\usepackage{microtype}

\usepackage{microtype}

\usepackage{graphicx}
\usepackage{amsfonts}
\usepackage{xcolor}
\usepackage{color}
\usepackage{xspace}
\usepackage{verbatim}

\title{Detecting Cross-Geographic Biases in Toxicity Modeling on Social Media}

\author{Sayan Ghosh \\
  University of Michigan \\
  \texttt{sayghosh@umich.edu} \\\And
  Dylan Baker \\
  Google Research \\
  \texttt{dylanbaker@google.com} \\\AND
  David Jurgens \\
  University of Michigan \\
  \texttt{jurgens@google.com} \\\And
  Vinodkumar Prabhakaran \\
  Google Research \\
  \texttt{vinodkpg@google.com} \\}

\date{}

\newcommand{\myparagraph}[1]{\paragraph{#1}}

\definecolor{c0color}{HTML}{5481A6}
\definecolor{c1color}{HTML}{7A117A}
\definecolor{c2color}{HTML}{2F7E2F}
\definecolor{c3color}{HTML}{C39B37}

\newcommand{\Czero}{{\textcolor{c0color}{\textbf{C0}}}\xspace}
\newcommand{\Cone}{{\textcolor{c1color}{\textbf{C1}}}\xspace}
\newcommand{\Ctwo}{{\textcolor{c2color}{\textbf{C2}}}\xspace}
\newcommand{\Cthree}{{\textcolor{c3color}{\textbf{C3}}}\xspace}

\newcommand{\fref}[1]{Figure~\ref{#1}}

\begin{document}
\maketitle
\begin{abstract}
Online social media platforms increasingly rely on Natural Language Processing (NLP) techniques to detect abusive content at scale in order to mitigate the harms it causes to their users. However, these techniques suffer from various sampling and association biases present in training data, often resulting in sub-par performance on content relevant to marginalized groups, potentially furthering disproportionate harms towards them. Studies on such biases so far have focused on only a handful of axes of disparities and subgroups that have annotations/lexicons available. Consequently, biases concerning non-Western contexts are largely ignored in the literature. In this paper, we introduce a weakly supervised method to robustly detect lexical biases in broader geocultural contexts. Through a case study on a publicly available toxicity detection model, we demonstrate that our method identifies salient groups of cross-geographic errors, and, in a follow up, demonstrate that these groupings reflect human judgments of offensive and inoffensive language in those geographic contexts. We also conduct analysis of a model trained on a dataset with ground truth labels to better understand these biases, and present preliminary mitigation experiments.

\end{abstract}

\section{Introduction}

Online social media platforms have increasingly turned to NLP techniques to detect offensive and abusive content online at scale (e.g., the Perspective API), as a way to mitigate the harms it causes to people.\footnote{\url{www.perspectiveapi.com}}
However, recent research has shown that these models often encode various societal biases against marginalized groups \cite{waseem2016you,caliskan2017semantics,park2018reducing,van2018challenges,dixon2018measuring,sap2019risk,davidson-etal-2019-racial,hutchinson-etal-2020-social,zhou2021challenges} and can be adversarially deceived \cite{hosseini2017deceiving,grondahl2018all,kurita2019towards}, potentially furthering disproportionate harm to those in the margins. 
These models are also challenged by several dimensions of noise, especially when compared with the corpora they are trained on. For instance, social media text across the globe may contain code switched language generated by multilingual speakers, neologisms, and other orthographic variations rarely seen in training data. This poses additional challenges to detect and mitigate fairness failures in these models.

\input{examples}

\begin{figure*}
  \centering
    \includegraphics[width=0.83\textwidth]{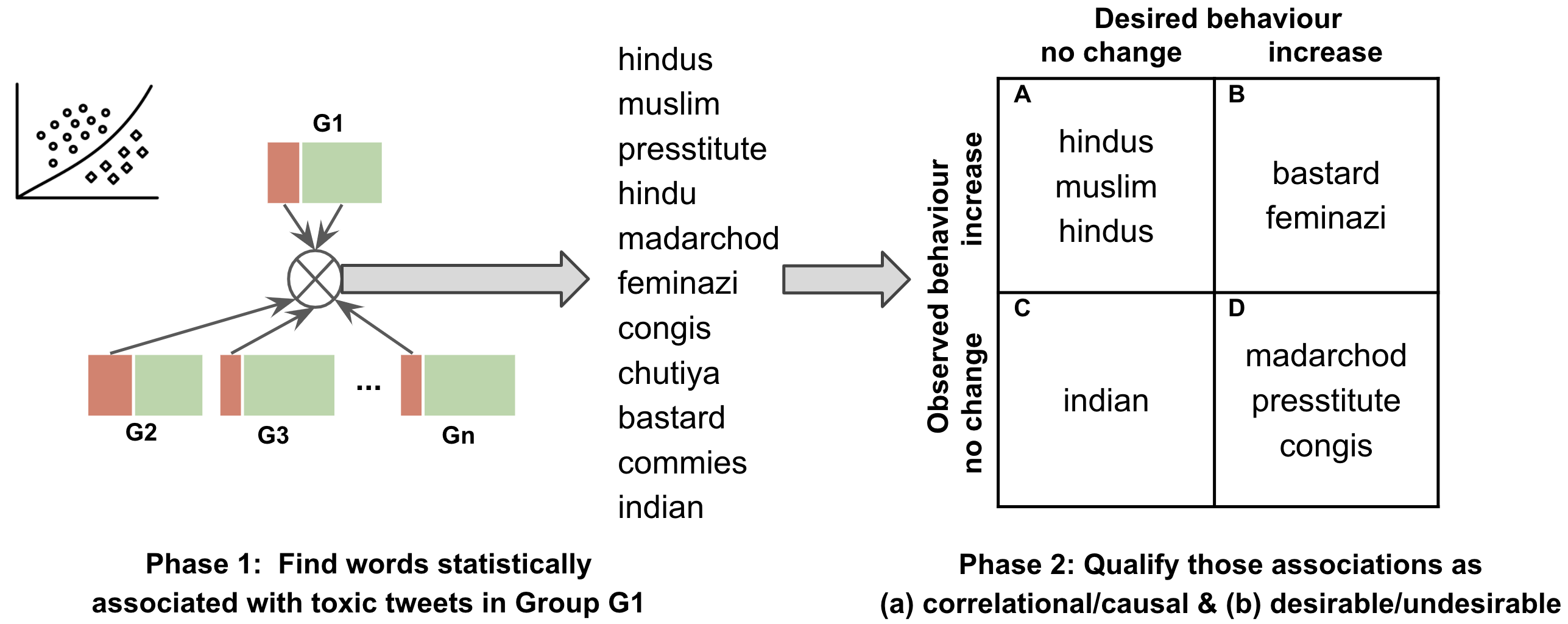}
  \caption{\small The high-level sketch of our two-phase methodology to identify undesirable model biases. \label{fig_methodology}}

\end{figure*}

In this paper, we propose a new weakly supervised method (outlined in \fref{fig_methodology}) to address shortcomings of current bias detection methods and robustly detect lexical biases in models (i.e. biases that associate the presence or lack of toxicity with certain words) when applied to noisy text (in particular, text generated by social media users from non-predominantly Anglophone nations). We draw upon the observation that social stereotypes and biases in data often reflect the public discourse at a specific point in time and place \cite{fiske2017prejudices,garg2018word}. 
We use a cross-corpora analysis across seven different countries to reveal terms overrepresented in each country. We then cluster the model behaviour against term perturbations to qualify these biases further.
Furthermore, we demonstrate the effectiveness of our method through human validation. Our method is task and genre agnostic; we expect it to work in any culture-laden tasks on genres of noisy text from platforms with global presence. Moreover, it can be applied in cases with and without ground truth labels. We conduct experiments on both cases and present a preliminary analysis on mitigation strategies to reduce harms created by these biases.

\section{Background}

Current research on detecting biases in NLP models primarily starts with a specific axis of injustice, and relies on annotated data \cite{sap2019risk,davidson-etal-2019-racial}
or lexicons that are salient to different subgroups. For example,  \newcite{dixon2018measuring} use a list of LGBTQ identity terms to identify varied outcomes in text classification and \newcite{hutchinson-etal-2020-social} use a list of terms that refer to persons with a disability. 
However, this reliance on lexicons and annotated data limit the focus to a handful of social injustices for which resources/lexicons are available, and largely ignores biases in non-Western contexts \cite{sambasivan2021re} and online environments containing creative or noisy language variation. Moreover, prior work shows that interpretation of hate varies significantly across different cultural contexts \cite{salminen2018online,salminen2019online} and reveals that the performance state-of-the-art models such as BERT \cite{devlin2019bert} on various tasks degrades significantly when used on noisy text \cite{sarkar2020non, kumar2020user}.

Consequently, current approaches to identifying model biases  miss 1) undesirable biases around words that are not captured in existing lexicons/datasets, and 2) desirable biases around offensive terms that are salient to different geocultural contexts that the models fail to capture. As an example, such biases are present even in a commonly-used toxicity detector like the Perspective API, as shown in Table~\ref{tab:examples}. 
In the first two sets of rows, we see that the mentions of certain identity groups (e.g., \textit{muslims}, \textit{Tamilian}) causes the toxicity model to assign a higher toxicity score. 
The third and fourth set of rows demonstrate that the model does not recognize the toxicity association on the words \textit{presstitute} and \textit{Madarchod},\footnote{\textit{Presstitute} is a portmanteau word blending the words \textit{press} and \textit{prostitute}, often used in Indian social media discourse; \textit{maadarchod} is an abusive word in Hindi, commonly used in Indian social media discourse in English.} whereas it does recognize the toxicity of words/phrases with the same/similar meanings. In this paper, we present a method to
detect such undesirable lexical biases.

\section{Methodology}
\label{sec_methodology}

We find undesirable lexical biases in two phases. The first phase aims to find words that are statistically overrepresented in tweets from specific countries that were deemed toxic by the model. 
Such overrepresentation could be due to different reasons: (1) the model has learned some of these words to be offensive and hence their presence \textit{causes} it to assign a higher toxicity score, or (2) some of the words appear more often in offensive contexts (e.g., frequent victims of toxicity) but their presence is merely \textit{correlational} (i.e., the model doesn't have any toxicity association for these words).
Another consideration is whether the association of a particular word towards toxicity is desirable or not. For some words, such as country-specific slur words, we want the model to be sensitive to their presence. On the other hand, we desire the models to not have biases towards culturally salient non-offensive words, such as religious concepts. 

The second phase attempts to separate the overrepresented words along these two dimensions:
\begin{itemize}[leftmargin=*,noitemsep,topsep=2pt]
    \item a descriptive axis 
    that separates words representing \textit{correlational} vs. \textit{causal} associations
    \item a prescriptive axis 
    that separates words representing \textit{desirable} vs. \textit{undesirable} associations
\end{itemize}
\noindent Both phases are described in Figure~\ref{fig_methodology} using tweets from India as an example. The \textit{causal-desirable} associations (e.g., \textit{bastard}) and \textit{correlational-undesirable} associations (e.g., \textit{indian}) capture the cases where the model is behaving as desired. On the other hand, the \textit{causal-undesirable} associations (e.g., \textit{hindus}) capture model biases that should be mitigated, whereas the \textit{correlational-desirable} associations (e.g., \textit{madarchod}) capture the geo-culturally salient offensive words the model missed. Our method aims to balance (i) achieving reasonably high precision and (ii) narrowing down the search space of words enough to be feasible in Phase 2 and any further human evaluation studies. We now discuss the methods we use for each phase.

\subsection{Phase 1: Identifying Biased Term Candidates}
First, for each country, we calculate the log-odds ratio with a Dirichlet informed prior \cite{monroe2008fightin} to find if term $i$ is statistically overrepresented in toxic tweets vs. non-toxic tweets in that country. 
Unlike the basic log-odds method, this method is robust against very rare and very common words as it accounts for a prior estimate of the expected frequency of each term. It also accounts for the variance in the word's frequency by calculating the z-score. 
However, common profanities are likely overrepresented in toxic tweets globally, so this initial list must be further filtered to identify geographically-specific terms. %
Since we now have a multi-class setting, one could repeat the \cite{monroe2008fightin} method in a one-vs-rest manner. However, such an approach would miss the important interdependence between these tests. 

Instead, we use the method from \citet{bamman2014gender} and \citet{chang2019automatically} that allows comparison across multiple groups of texts without requiring separate binary comparisons.
We consider the term $i$ is overrepresented in toxic tweets from a country $j$ if $i$ occurs with higher than statistically expected frequency in that country. Similar to prior works, we assume a non-informative prior on $f_i$, with a $\mathrm{Beta}(k_i, N-k_i)$ posterior distribution, where $k_i$ is the count of $i$ in the geography-balanced corpus and $N$ is the word count of the corpus.
We use a balanced corpus with the same number of toxic tweets from each country (matching the one with the least number of tweets). Term $i$ is deemed significantly associated with country $j$ if the cumulative distribution at $k_{ij}$ (the count of term $i$ in the corpus corresponding to country $j$) is $\leq 0.05$. 

By combining these two methods, we find culturally salient words that are overrepresented in toxic tweets in each country. 
We also conducted experiments using \cite{monroe2008fightin} method in a one-vs-rest manner (described above), as well as using \cite{bamman2014gender} method for both steps; but they provided qualitatively worse results than the above approach.

\input{table_words}

\subsection{Phase 2: Qualifying Associations}

To determine the causal vs. correlational distinction, one could rely on counterfactual approaches such as perturbation sensitivity analysis (PSA) \cite{prabhakaran2019perturbation}. However, PSA (and similar approaches) aggregate the sensitivity to terms
by calculating the mean difference in prediction scores 
across all sentences. Here, we investigate how the sensitivity varies across the range of toxicity scores, which gives important clues about the desirability of biases. 
Using naturally occurring sentences like \cite{prabhakaran2019perturbation} would introduce numerous unknown correlational effects. Hence, we followed \citet{hutchinson-etal-2020-social} and \citet{may2019measuring} by 
building a set of $33$ template sentences (see Appendix~\ref{app:templates}) that correspond to a range of toxicity scores.
A few examples below:
\begin{itemize}[noitemsep,topsep=5pt] 
    \item You are a \textit{person}
    \item I really hate it when \textit{person} is there
    \item I really dislike \textit{person}
    \item I am going to the movies with a \textit{person}
    \item The \textit{person} was going to do that with me
\end{itemize}
\noindent For each term obtained in Phase 1, we measure the shift in toxicity in response to replacing the word \emph{person} in the template sentence with that term.
 
\begin{figure}
    \centering
   
    \begin{center}
    \includegraphics[width=0.43\textwidth]{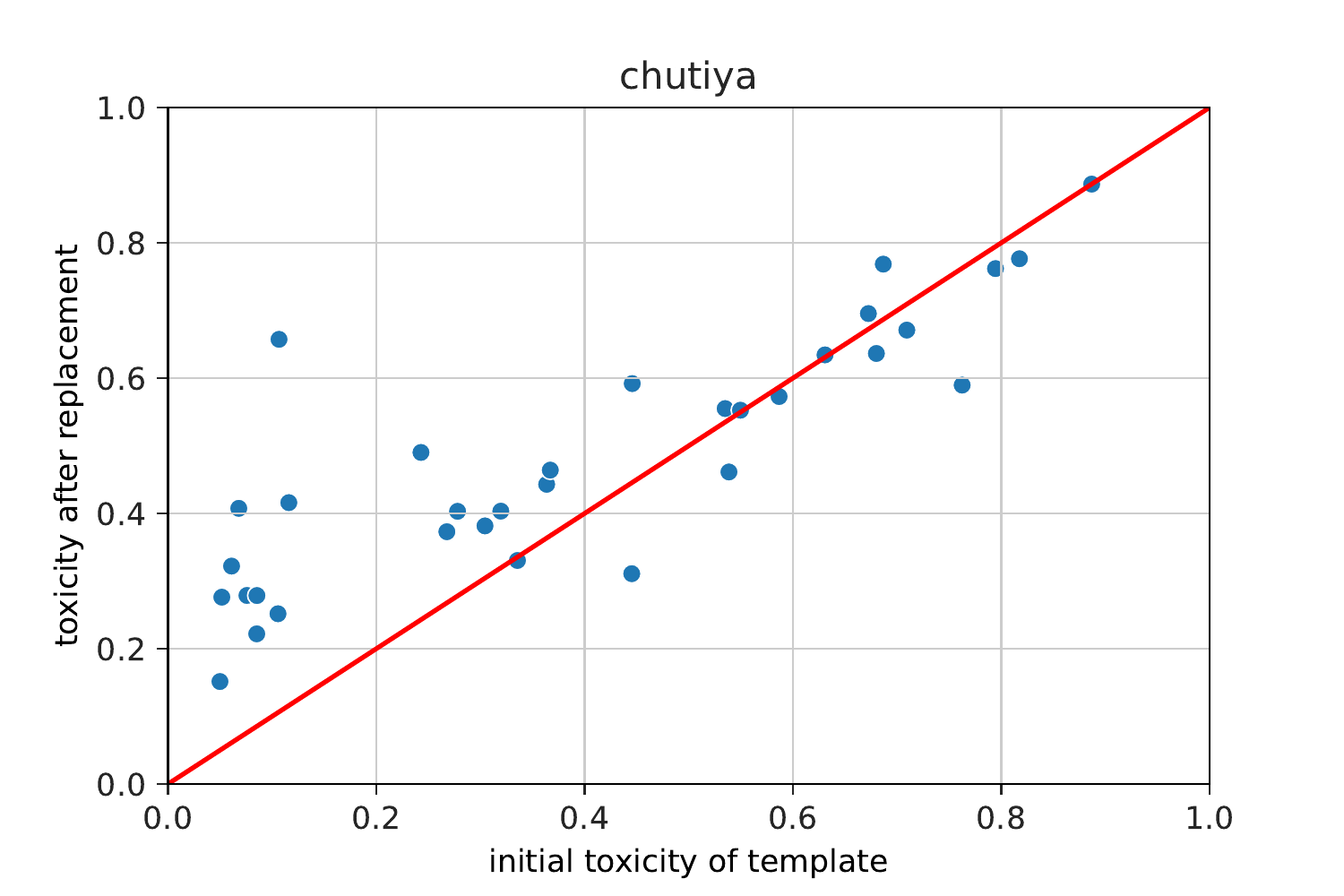}
    \caption{Example of vectorized deviations used in clustering for the term ``chutiya". The $x$-axis indicates the original template scores. The $y$-axis represents the score of the perturbed template. 
    }
    \label{fig:cluster-example}
    \end{center}
\end{figure}

Ideally, words with little to no inherent offensiveness should have no effect on a model's estimate. Words that are highly offensive should increase the score across the board. Deviations from these may indicate either undesirable biases or knowledge gaps in the model. To model different patterns of behavior by the classifier, we construct vector representations for each term by creating vectors $\mathbf{x} \in \mathbb{R}^d$  such that $x_i$ is the toxicity after replacing \emph{person} in the $i$-th template with a word from a lexicon. An example of a visual representation of these vectors is shown in  \fref{fig:cluster-example}. We then cluster terms that have similar behaviour together to qualitatively distinguish the kind of association they have with toxicity.
In this paper, we use the $k$-means clustering algorithm; however, our approach is agnostic to what clustering method is used. 

\section{Evaluation of Off-the-shelf Models}

Our approach can notably be applied to cases where ground truth labels are not available. This is useful in deployment settings, where it may not be feasible and/or be expensive to annotate sufficient amounts of data regularly or where one does not have access to the training data but wants to understand biased outcomes. We provide an example of how our method could be used in this setting. 

\subsection{Experimental Setting}

We use the Perspective API's toxicity model for our analysis. We use 73 million tweets from across 7 countries with substantial English-speaking populations that are active online: India, Pakistan, Nigeria, Ghana, Jamaica, Mexico, and the Philippines. We focus on these countries as opposed to countries that are solely and primarily Anglophone as to better understand how models interact with ``non-standard" English dialects, especially ones that are different from the data the model was likely trained on. Our data is collected from a $\sim$10\% random sample of the tweets from 2018 and 2019, from the Decahose stream provided by Twitter.  We pre-process tweets by removing URLs, hashtags, special characters and numbers, and applying uniform casing. We also lemmatize each resulting word to control for factors such as plurality using NLTK \cite{loper2002nltk}.
The number of tweets from each country is shown in Table~\ref{tab:example-words}.

\subsection{Results}

\myparagraph{Phase 1:}
Our Phase 1 analysis
identified several hundred terms overrepresented in each country. The number of terms as well as a small subset of terms for each country are shown in Table~\ref{tab:example-words}. As evident, our method is effective at picking up words that are geography-specific, compared to simply using log-odds to identify overrepresentation in toxic versus non-toxic tweets; such an approach outputs more general, widely used profanities that (by nature) occur more commonly in offensive messages across all countries.
Crucially, not all of the words identified by our method are inherently toxic: For example, the list for Pakistan contains words such as \textit{muslims} and \textit{journalism}, which should not carry a toxic connotation and could be reflective of model biases.
Further, our method identifies emergent country-specific terms such as \textit{presstitutes} (a slur used to refer to members of the press) and \textit{congi} (a slur for supporters of the Congress Party, a large political party) in India.

\begin{figure*}[t]
    \centering
    \begin{subfigure}{0.3\textwidth}
        \includegraphics[width=\textwidth]{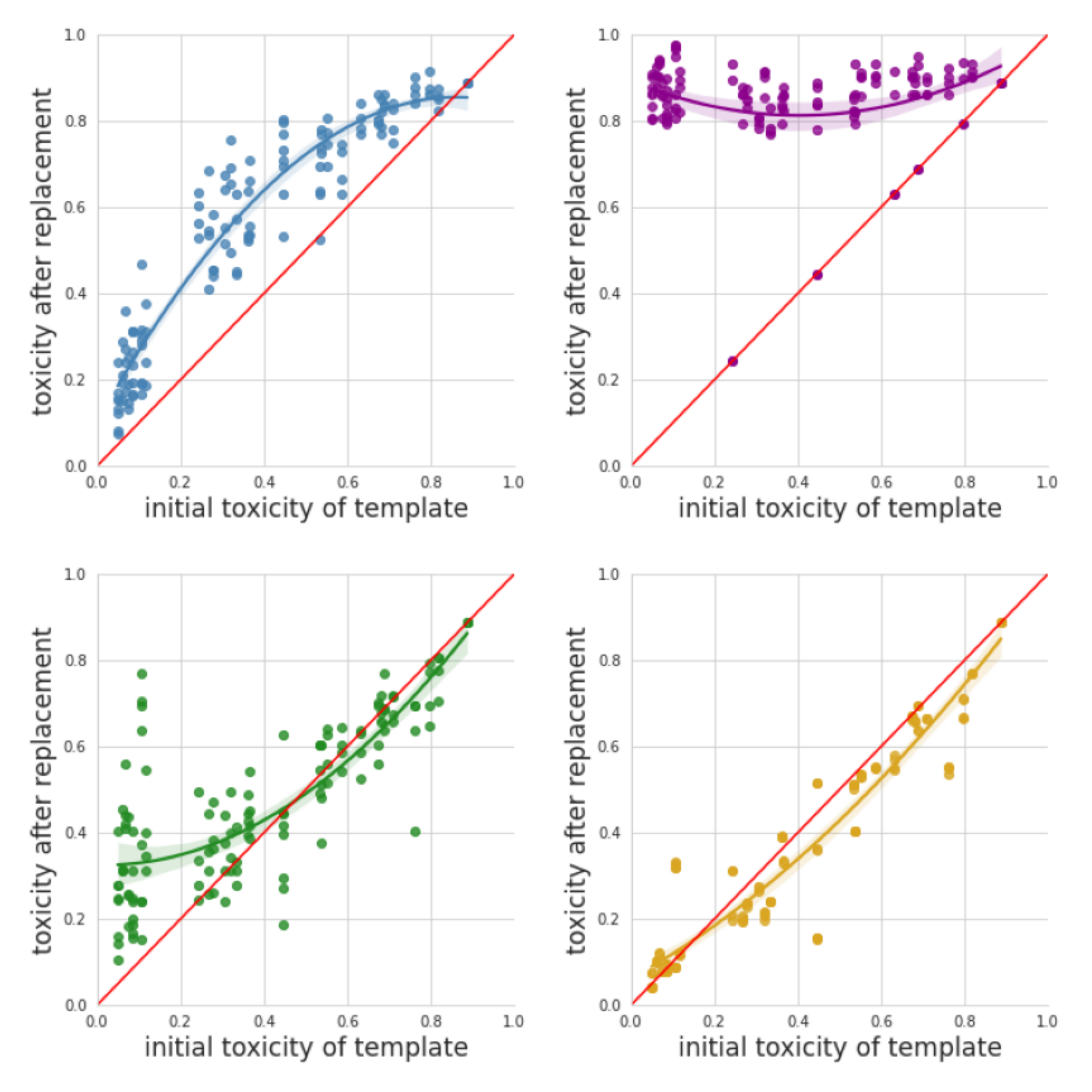}
        \caption{India}
    \end{subfigure}
    \hfill    
    \begin{subfigure}{0.3\textwidth}
        \includegraphics[width=\textwidth]{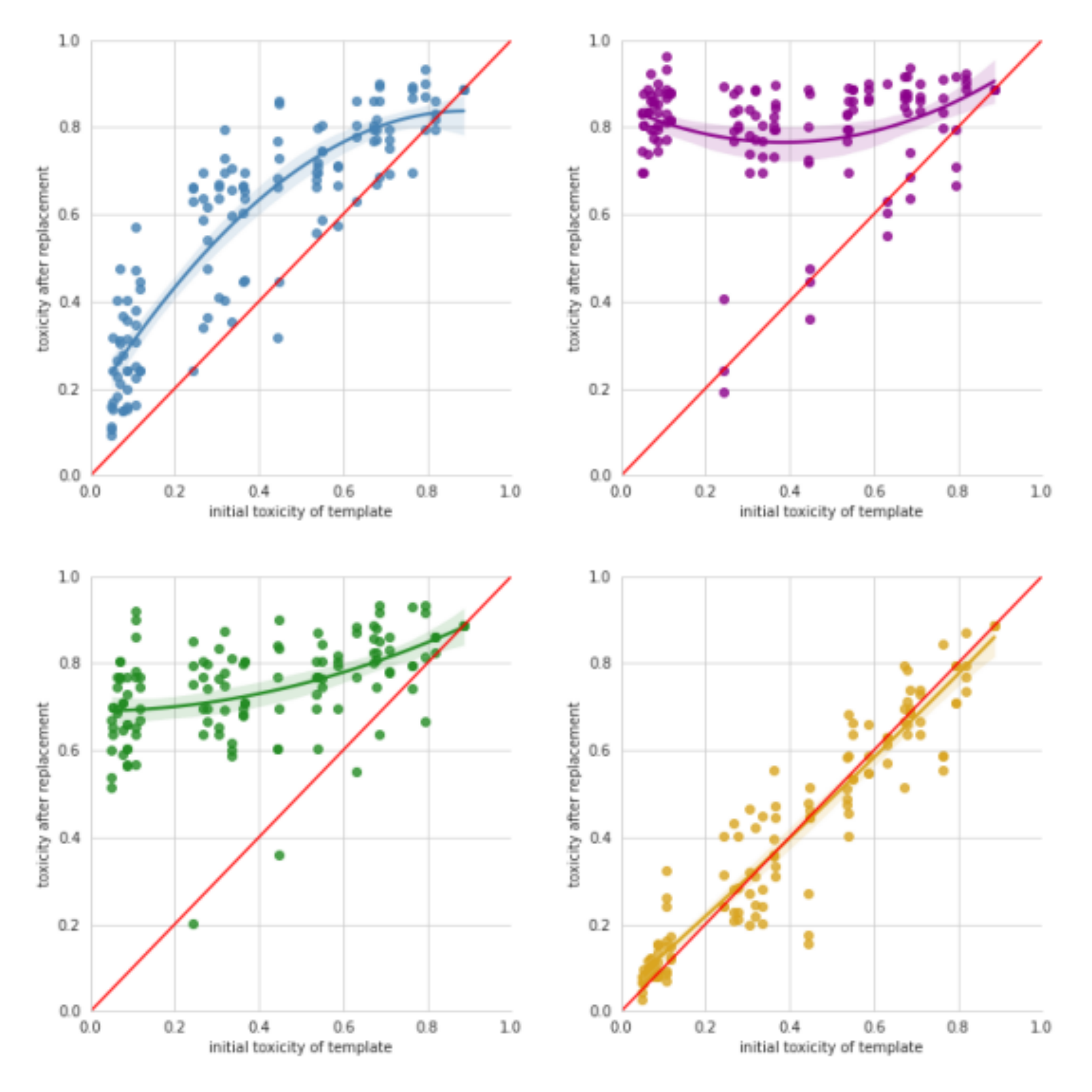}
        \caption{ Mexico }
    \end{subfigure}
    \hfill
    \begin{subfigure}{0.3\textwidth}
        \includegraphics[width=\textwidth]{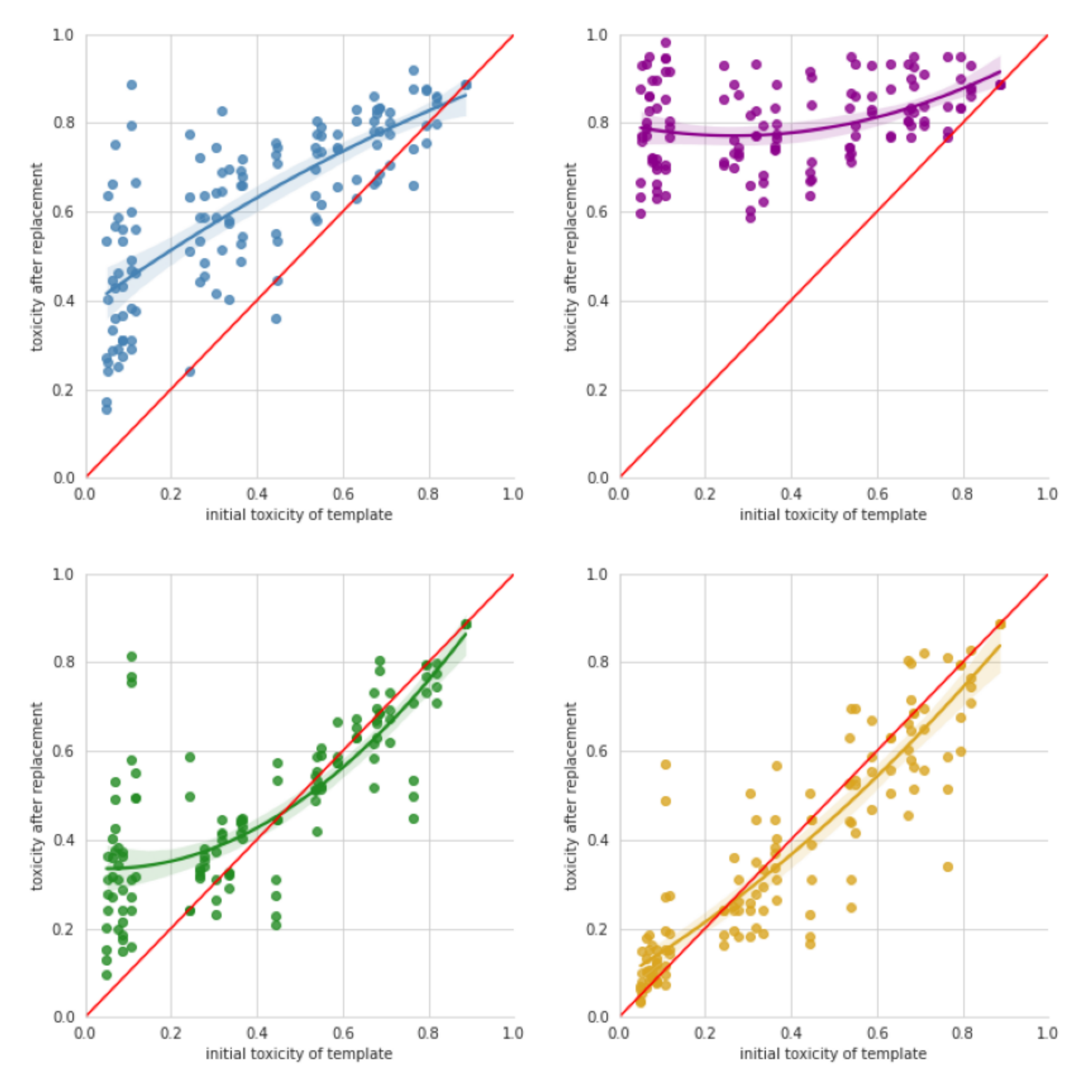}
        \caption{ Nigeria }
    \end{subfigure}
    \hfill        
    \label{fig:first}   
    \caption{Changes in offensiveness scores of template sentences in response to the use of over-represented terms for each cluster, for India, Mexico, and Nigeria. 
    }
    \label{fig:clusters}
\end{figure*}

\myparagraph{Phase 2:} 
In Phase 2, we analyze the model behaviour in response to perturbations using each term from Phase 1, and cluster them based on model sensitivity. Here, we focus on nouns and named entities which represent $77 \%$ of the terms from phase $1$ and are the most salient to analyze in terms of subgroups (although the method can be altered to use templates that allow for the use of other parts of speech). We use $k$-means clustering; we selected $k=4$ based on manual inspection of $k=[3,6]$; this inspection showed that the majority of instances fall into four natural clusters, with lower $k$ merging dissimilar instances or higher $k$ producing clusters with very few instances.
Figure~\ref{fig:clusters} shows the model behaviour of each cluster we obtained for India, Mexico, and Nigeria; figures for the remaining countries are shown in Appendix~\ref{app:other-country-results}, with a sample of words in each cluster. 
We list a sample of words in each cluster for India below:

\begin{itemize}[leftmargin=*,noitemsep,topsep=5pt] 
\item \textbf{\Czero}: \input{cluster1}
\item \textbf{\Cone}: \input{cluster2}
\item \textbf{\Ctwo}: \input{cluster3}
\item \textbf{\Cthree}: \input{cluster4}
\end{itemize}

\noindent Qualitative inspection of the clusters reveals certain properties of the words within each.
Cluster \Czero (top left of Figure~\ref{fig:clusters}a) contains words such as \textit{feminist} and \textit{muslim} that increase the toxicity of the templates with the largest changes occurring in the middle, indicating undesirable biases in the model.
In contrast, cluster \Cone (top right) has words such as \textit{shithead} that 
uniformly raise the toxicity of the templates to the very high range, regardless of the initial toxicity of the template.

Words in cluster \Ctwo (bottom left) display a similar early trajectory to those in cluster \Czero, but converges towards the diagonal. Words in this cluster include \textit{presstitute} and \textit{porki}, both of which are country-specific slurs, suggesting that the model only has a weak signal about their toxicities.
Similarly, cluster \Cthree (bottom right) consists of words that do not affect the templates much at all after replacement. Some words in this cluster are indeed not toxic, but we also see words such as \textit{congi} that the model has not encountered before, but carry a negative connotation in Indian online discourse.

\begin{figure*}[t]
    \centering
    \begin{subfigure}{0.3\textwidth}
        \includegraphics[width=\textwidth]{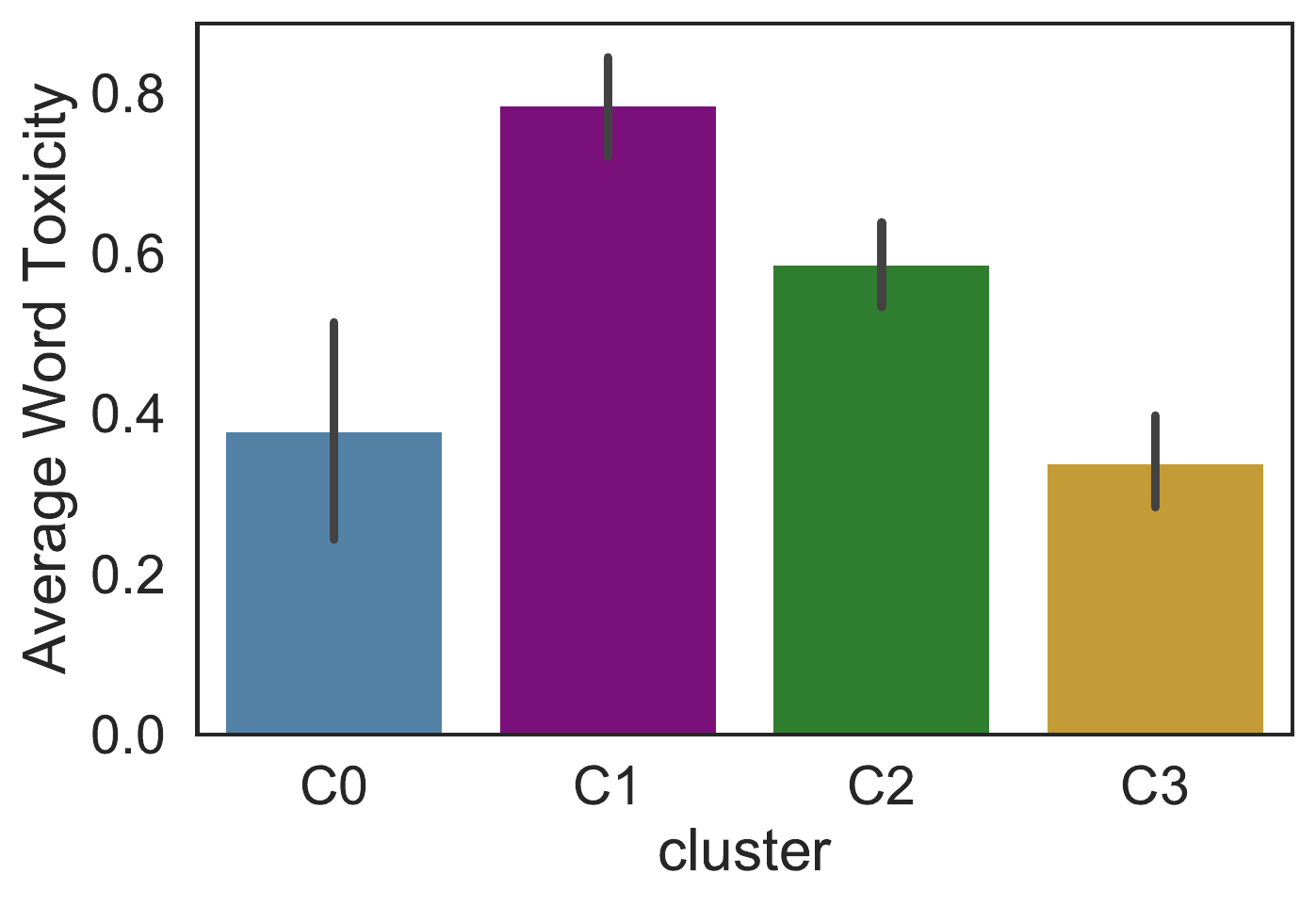}
        \caption{India}
    \end{subfigure}
    \hfill
    \begin{subfigure}{0.3\textwidth}
        \includegraphics[width=\textwidth]{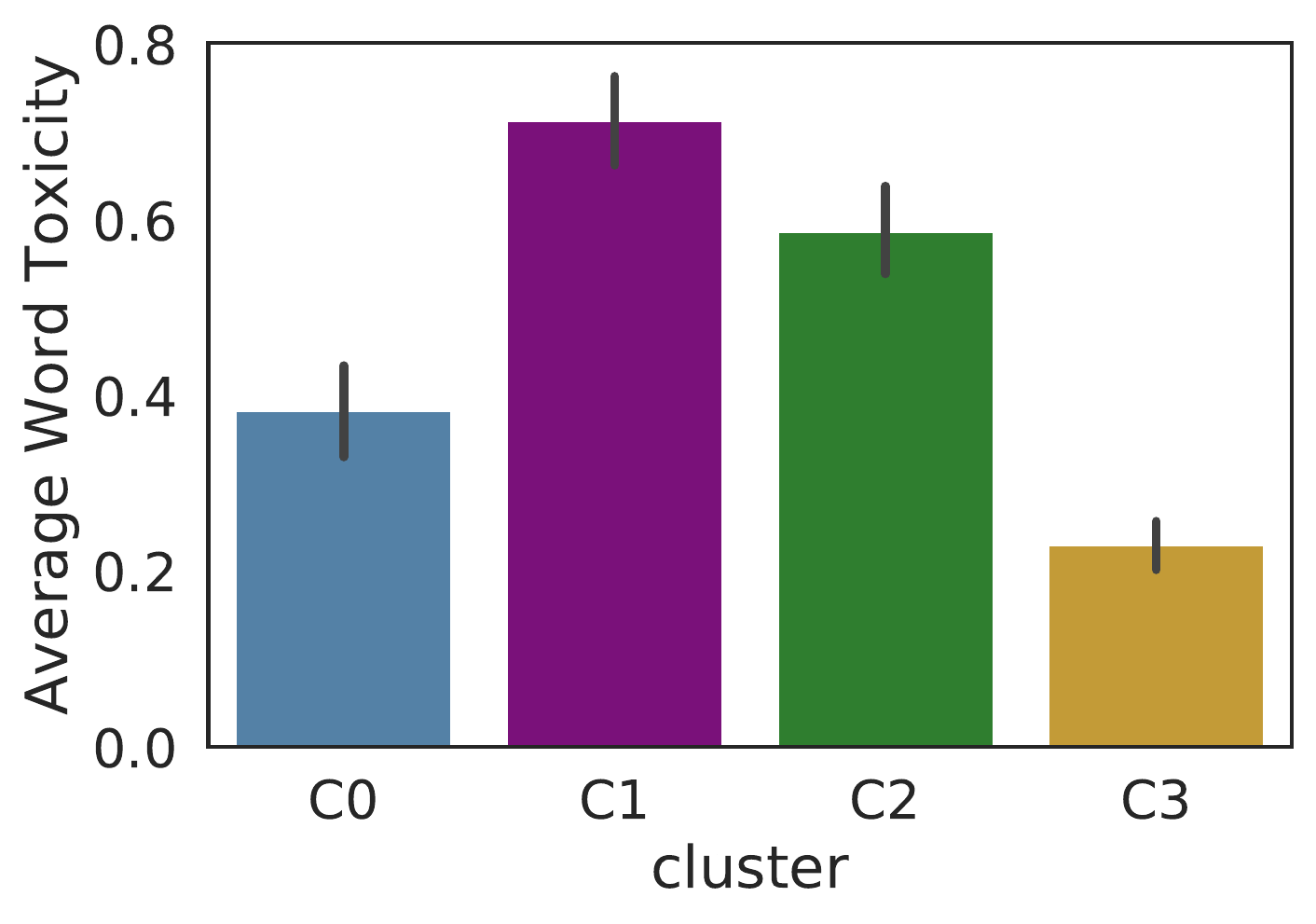}
        \caption{Mexico}
    \end{subfigure}    
    \hfill    
    \begin{subfigure}{0.3\textwidth}
        \includegraphics[width=\textwidth]{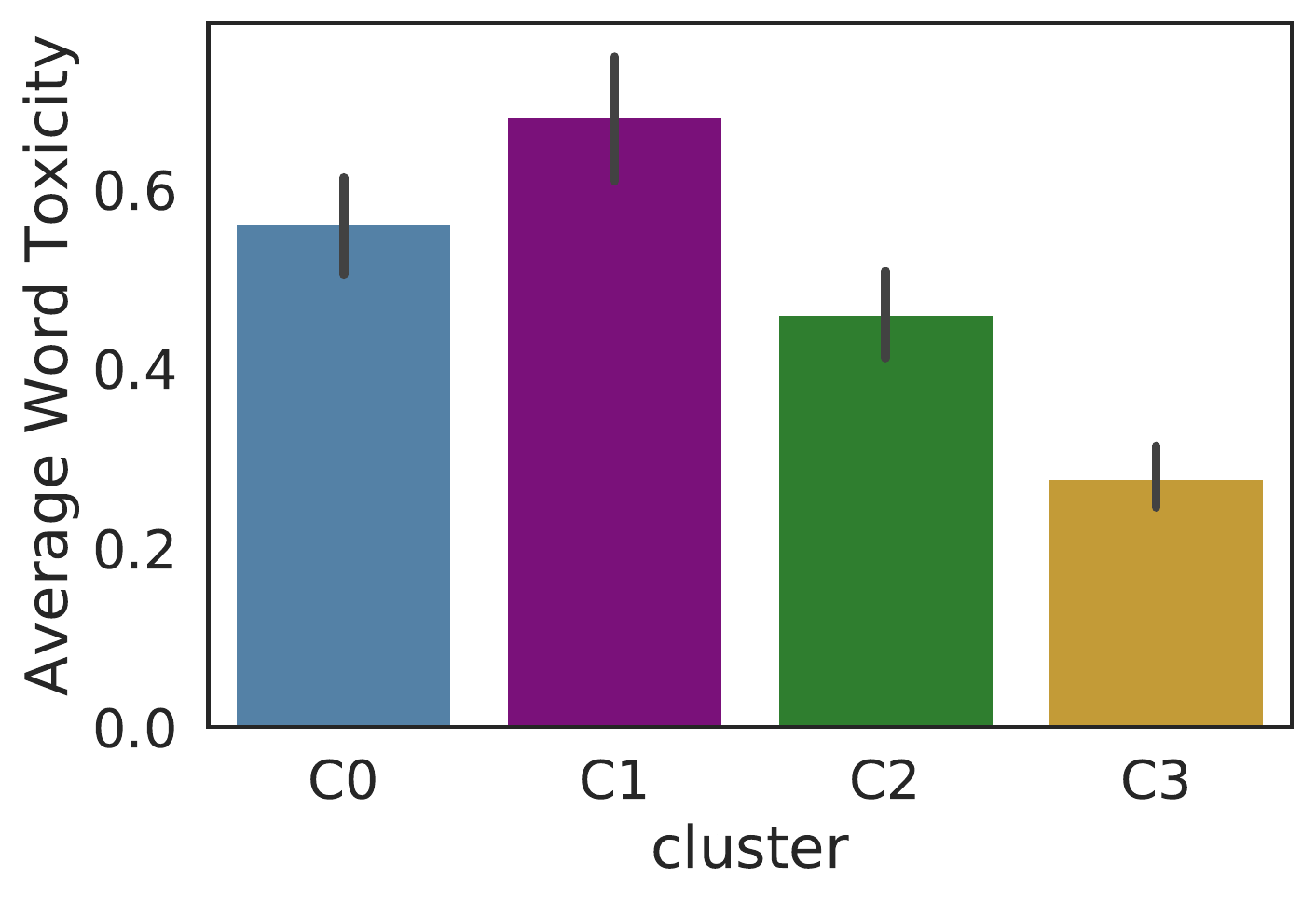}
        \caption{ Nigeria }
    \end{subfigure}
    \caption{
        In-community judgments of word offensiveness by cluster, shown here for India, Mexico, and Nigeria (cf. \fref{fig:clusters}) show that the method is indeed finding biases perceived by in-community speakers.}
    \label{fig:qualtrics}
\end{figure*}

\subsection{In-Community Validation}

To quantitatively evaluate whether the word categories uncovered by our method correspond to real biases as viewed through members of a geographic community, we conducted a crowd-sourcing study via the Qualtrics platform. We note that even despite this step, there may be certain forms of bias that are not uncovered by our method. However, this form of validation attempts to judge whether the biases that our method does uncover have a correspondence with real-world judgments. 

We recruited 25 raters per country, and each was asked to rate 60 terms (including 10 control words) to be (i) inoffensive, (ii) sometimes offensive, (iii) highly offensive, or (iv) don't know this word. We obtained 5 ratings each for 250 words per country. We removed ratings from raters who got more than 3 out of 10 control words wrong. Full annotation details are provided in  Appendix~\ref{app:in-community-analysis}.\footnote{
Despite recruiting a large pool of annotators for each instance, our annotators may not entirely capture the value systems of countries due to selection effects of who has access to the internet and is able to participate in Qualtrics work. For instance, raters from India in our human ratings  potentially overrepresent views of the middle class, those of upper caste individuals, and views from a handful of states with tech hubs. As a result, their perception of what is or is-not offensive may not be representative of India at large.} 

\looseness=-1
We hypothesize that the words grouped into \Czero would be judged identical to those in \Cthree (i.e., treated as false positive features), where \Ctwo should be more offensive, potentially  up to \Cone (i.e., false negative). 
In aggregate, the annotator ratings largely matched our hypotheses across countries. 
\fref{fig:qualtrics} shows the ratings for the clusters from India, Mexico and Nigeria (cf. \fref{fig:clusters}); here, we re-scale their categorical ratings into [0,1]. Our results show that the words in \Czero and \Cthree  are often very similar in their perceived toxicity and, in the case of India, have no statistical difference in their toxicity ratings. Our analysis on Nigerian tweets does differ in this trend, however, where \Czero is perceived to be more toxic than \Cthree---and even slightly more than \Ctwo. Words in this cluster include \textit{feminist}, \textit{junkie}, \textit{africans}, and \textit{maggot}, which, though slightly negative in connotation, would not appear to be overly offensive and in-line with expectations for what \Czero should contain. This difference for Nigeria in our case study  exemplifies how in-community interpretations do differ from those of external readers and underscores the importance of including these viewpoints when assessing model errors.

\looseness=-1
The \Ctwo clusters, which contains words like ``presstitute'' in India, were found offensive, matching our hypothesis of these words being false negatives. In all countries, the level of offensiveness was still less than the level for extremely toxic words in \Cone, but still larger than \Cthree.Figures for other countries are shown in the Appendix~\ref{app:other-country-results}. 
The clusters found by our method do correspond to in-community judgments of toxicity and reflect meaningful groups of biases in the model, which can be used by experimenters for later bias testing and mitigation.

\section{Towards Mitigating Undesirable Biases}

Our pipeline provides an unsupervised approach to identifying potential latent biases in toxicity models. Ideally, practitioners aim not only to identify these biases but also to \textit{fix} them before deploying their systems. However, %
our method identifies terms of different types: causal-desirable associations and correlational-undesirable associations, as well as causal-undesirable and correlational-desirable associations both of which require different approaches to handle. The clusters we find in Phase 2 approximate these different types of associations, and hence we expect them to behave differently for different mitigation approaches. Here, we test the impact of baseline  mitigation techniques on reducing bias in these clusters.

\subsection{Experimental Setting}

Performing bias mitigation experiments involves re-training the model, and hence requires toxicity annotated training data. Since we do not have access to the training data used to train the Perspective API model, we employ the dataset provided by Jigsaw for the Kaggle competition on toxicity classification \cite{Jigsaw-bias}
that contains public comments from the Civil Comments platform and annotated toxicity ratings (on a continuous scale in the range $[0, 1]$). Here, as per the original guidelines provided by Jigsaw, we consider an instance with a rating of $>= 0.5$ as toxic. We train the toxicity model for this analysis by fine-tuning a DistilBert model \cite{sanh2019distilbert} using pre-trained parameters from HuggingFace on the Jigsaw dataset.
The model is fine-tuned for one epoch with an initial learning rate setting of $2 \times 10^{-5}$.\footnote{The model was trained for three epochs with early stopping; ultimate model performance on a held-out 10\% development set was highest after one epoch.}

While some performance differences are inherent between this model and the Perspective API model, our method still identifies similar latent biases. 
Phase 1 of our pipeline identifies 150 candidate words with bias. These include words such as \textit{liberal}, \textit{muslim}, \textit{sanghi}, and \textit{commie}. In particular, we also note that the set of words flagged using the scores from our model is a strict subset of the words flagged using the scores from Perspective API. To quantify the effects of mitigation strategies, we further analyze 120 of these words that appear in at least 10 instances each for the toxic and non-toxic labels in the Jigsaw dataset.

\subsection{Bias Measurement}

\begin{table*}[t]
\resizebox{\textwidth}{!}{
\centering
\begin{tabular}{@{}lcccccccc@{}}
                         & \multicolumn{6}{c}{AUC/AEG Bias Metrics (Mean)}                   &     \multicolumn{2}{c}{Toxicity Performance}         \\
                         \cmidrule{2-6} \cmidrule{8-9}
\textbf{Model}                    & \textbf{Subgroup-AUC} $\uparrow$ & \textbf{BPSN-AUC} $\uparrow$ & \textbf{BNSP-AUC} $\uparrow$ & \textbf{AEG+} $\downarrow$ & \textbf{AEG-} $\downarrow$ && \textbf{F1} &\textbf{ Overall AUC} \\\midrule
No Mitigation      & 0.91              & 0.88          & 0.98    & 0.08 & 0.25      && 0.70      & 0.97        \\
Substitution             & 0.89              & 0.86          & 0.96 & 0.07 & 0.26         && 0.63      &         0.96    \\
Deletion                 & 0.88              & 0.86         & 0.96 & 0.07 & 0.26          && 0.60      &        0.95     \\
Balance and Tune ($k$=100)   & 0.91              & 0.90         & 0.97 & 0.07 & 0.24         && 0.69      & 0.97         \\
\end{tabular}
}
\caption{Model performance after applying bias-correction metrics, showing that the mitigation was largely ineffective at reducing bias. Arrows indicate the direction of less model bias. }
\label{tab:debiasing-results}
\end{table*}

As \newcite{borkan2019nuanced} note, biases in models can be nuanced and varied, and require multiple metrics that can provide insights on how to mitigate them. %
They compute a suite of five metrics based on subgroups of instances defined around the presence of different identity terms. We adopt a similar approach by defining 120 subgroups based on the presence of each of the 120 terms our bias evaluation pipeline identified. 

The first three metrics are based on AUC. 
(1) Subgroup AUC
reflects how separable are the toxic/non-toxic instances containing our target words. 
(2) Background Positive Subgroup Negative (BPSN) AUC measures how separable are non-toxic instances containing the target words from toxic instances without them; lower scores are typical of models with false positives.
(3) Background Negative Subgroup Positive (BNSP) AUC measures how separable are non-toxic instances without the target words from toxic instances with them; lower scores are typical of models with many false negatives.
As AUC, these scores also range $[0,1]$ and higher scores are desirable.

The last two metrics compare the toxicity scoring distributions of the instances with the target words and those instances without. These metrics measure the Average Equality Gap (AEG) for toxic and non-toxic instances (separately) within $[-0.5, 0.5]$, where scores close to 0 indicate no bias in how instances with the target word are treated. The AEG metrics capture cases where the scoring distributions of the target words are shifted but not sufficiently to cause re-orderings (by toxciity) that would be identified by the AUC metrics. We refer the reader to \citet[][\S3.3]{borkan2019nuanced} for a detailed discussion of what these metrics capture.

\subsection{Mitigation Approaches}

Mitigating undesirable biases in NLP models is an active research area \cite{zhou2021challenges}, where counterfactual and data augmentation techniques tend to be the most common approach \cite{dixon2018measuring,garg2019counterfactual,davani2020fair}. In this section, we present experiments using three such , with the goal to demonstrate that different approaches are required to mitigate different types of biased associations our pipeline reveals.

\textbf{Deletion}: Here, we delete all training instances containing a biased word and re-do the fine-tuning of the DistillBERT model using the remaining instances. In doing this, we expect to remove undesirable associations around the target words learned from the training data, however it will remove a lot of useful information as a side effect. 

\textbf{Substitution}: Here, we substitute all occurrences of the target words in the training data with a \texttt{<UNK>} token. This is similar to the \textit{blindness} setting used by \cite{garg2019counterfactual}. Unlike the Deletion approach, this approach aims to remove undesirable associations without the side effect of dropping useful instances.

\textbf{Balance and Tune}: Finally, we use an approach that further fine-tunes the model using a subset of the dataset where each target word has an equal number of toxic and non-toxic instances. Formally, for $m$ occurrences of the target word in instances with the toxic label and $n$ instances with the non-toxic, we select $max(k, min(m,n))$ instances of each and fine-tune the model further only on these instances. 
We experiment with $k$=10, 50, 100. For all three experiments, we use the same fine-tuning hyperparameters as in the baseline model.\footnote{Separate hyperparameter tuning on the held-out development set showed these parameters to still be optimal.}

\begin{figure*}[t]
  \centering
    \includegraphics[width=\textwidth]{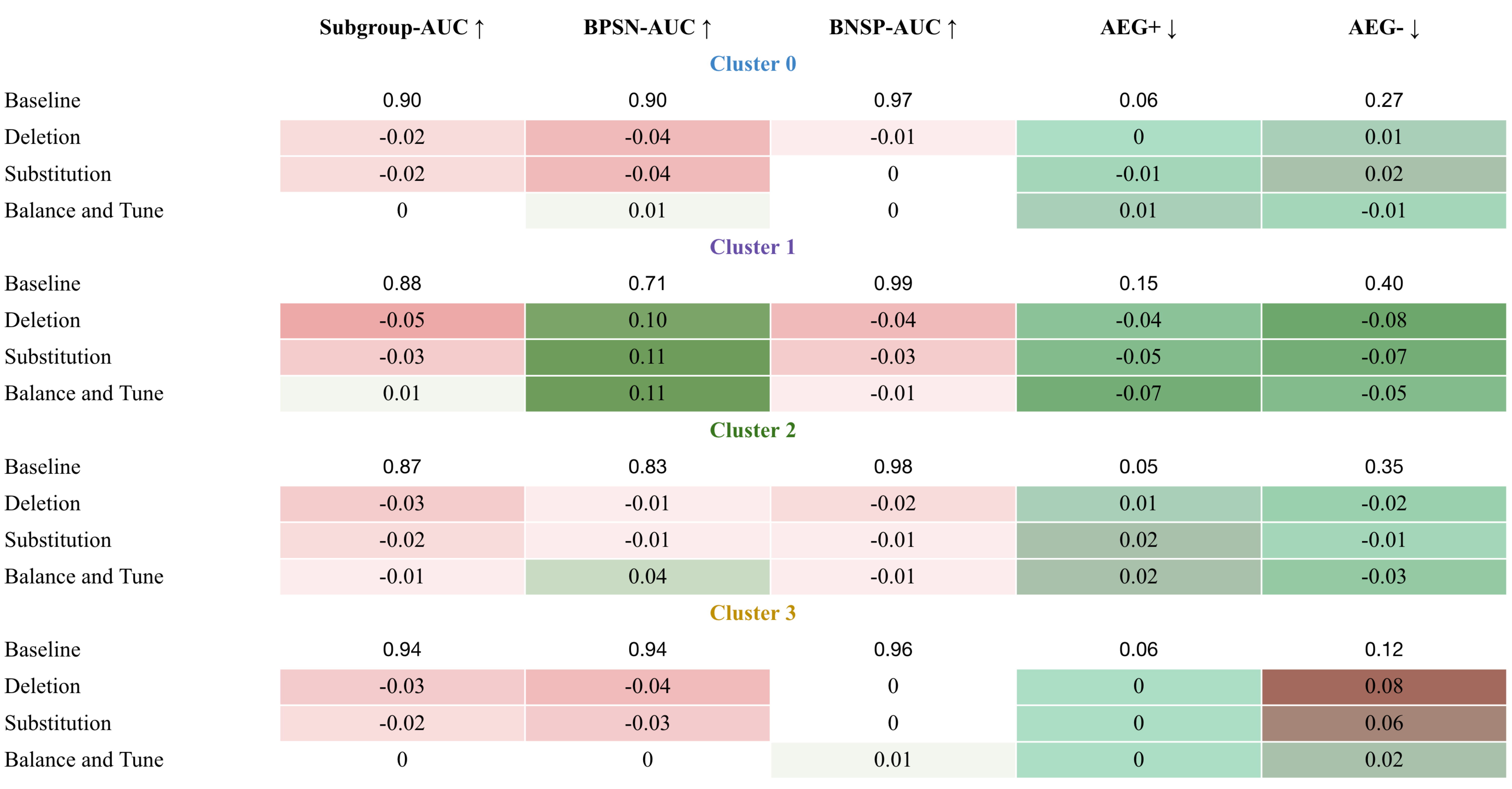}
  \caption{\small Cluster-level disaggregated bias mitigation results. . \label{fig_cluster_results}}
\end{figure*}

\subsection{Results}

In aggregate, none of the mitigation strategies substantially reduced the bias, as shown in Table \ref{tab:debiasing-results}.\footnote{Results for the Balance and Tune models were similar for all $k$. We show results with other $k$ values in Appendix~\ref{app:additional-mitigation}.}
Three notable trends were observed.
First, mitigation strategies varied in their net effect on recognizing toxicity. Substituting the biased word for \texttt{<unk>} or deleting those instances entirely had a detrimental effect on model F1. In these cases, AUC remained largely the same as without mitigation suggesting these models are still capable of ordering instances correctly by toxicity, but the overall scoring has changed sufficiently to affect classification decisions. In contrast, the Balance and Tune had little effect on classifier performance.

Second, the Substitution and Deletion models both \textit{increase} bias on every metric except AEG+. Despite not training models on any instances containing the target terms, the resulting models still produced classification decisions biased against the target words. This result suggest that the pre-trained DistilBERT model itself is likely responsible for the latent bias, mirroring multiple results showing biases in these models \cite{kurita2019measuring, shwartz2020you, nadeem2020stereoset}.

Third, the Balance and Tune strategy had negligible effect on both bias correction or toxicity performance. This result suggests that starting from a biased classifier based on a pre-trained language model, the bias cannot easily be corrected by simply rebalancing the class distribution in training data and further fine-tuning.  Together, our results point to how simple but seemingly common-sense approaches  fail to substantially reduce bias and thus motivate the need for (i) more sophisticated mitigation strategies or (ii) pre-trained language models free of such latent biases.

Do bias mitigation approaches change performance differently for each cluster of terms? Figure~\ref{fig_cluster_results} shows cluster-level disaggregated bias metrics for the baseline model, as well as the increases/decreases obtained through the different mitigation approaches, showing that the bias metrics do vary across different clusters.
The high AEG+/- of clusters \Czero and \Ctwo indicates that both clusters, on average, score higher than background examples, confirming the intuitive notions captured by the visual examination of the clusters (i.e. the terms in these clusters tend to be rated higher than their corresponding ``neutral" templates). Cluster \Cone also has higher AEG metrics, but these are mostly toxic words that the model already recognizes. 
On the other hand, \Cone and \Ctwo fares poorer than the other two on AUC bias metrics, especially in BPSN-AUC. 

Our bias mitigation approaches also act differently on different clusters. Overall, the Balance and Tune technique has the most gains in bias metrics in \Cone, and moderate gains in \Ctwo, whereas it has negligible impact on \Czero and \Cthree. On the other hand, Deletion and Substitution approaches significantly worsened AEG metrics of \Cthree, as well as AUC metrics, across the board except for BPSN-AUC of \Cone. These results demonstrate the need for different approaches to mitigate the different kinds of  biased associations our pipeline reveals. However, more work needs to be done on finding which mitigation approaches are appropriate for different kinds of undesirable biases.

\section{Discussion and Conclusion}

Toxicity detection models commonly rely on training data that reflects discourse that is temporally and culturally situated. As such, models may contain biases by learning undesirable  features that are not toxic or be blind to toxic language underrepresented in their training data. Here, we introduce a new weakly supervised approach for uncovering these biases.\footnote{All code and data to replicate the results will be released at the url: \url{https://github.com/sghosh73/Cross-Cultural-Bias-Detection}}
Our method provides a key diagnostic tool for model creators and deployers to test for biases. From a fairness perspective, our analysis reveals biases present in the model along different dimensions, without having any labeled resources for those dimensions. For instance, \Czero revealed biases around religion (\textit{muslim}, \textit{hindu}, \textit{jews}), country/ethnicity (\textit{indian}, \textit{pakistani}, \textit{arabs}), and ideology (\textit{feminist}, \textit{liberals}, \textit{communist}) without looking for these axes a priori. This could be a valuable first step before deeper analysis of these biases.

At a practical level, our method reveals country-specific offensive terms that the model has not seen before (e.g., \textit{maadarchod}). Such country-specific abusive language lexicons could aid NLP practitioners trying to make their models robust across geographies. 
Such lexicons could also aid human-rights organizations working on building region-specific hate-speech lexicons.\footnote{\url{https://www.peacetechlab.org/}}
Furthermore, our model also identifies portmanteau (blended) words that are used disproportionately in certain geo-cultural contexts (e.g., \textit{presstitute}, and \textit{libtard} in Indian social media discourse). Our method could help monitor emergent offensive language in different geographies and aid data augmentation efforts in keeping models up-to-date through surfacing emergent language in new geographies.

\section{Ethical Considerations}

The proposed method provides an efficient distantly-supervised method for practitioners to identify potential biases in their toxicity detection methods. Although intended strictly as beneficial, it could create the risk of overconfidence in a lack of bias by a particular model. While we have demonstrated that the approach identifies clusters of words that mirror  in-group judgments of (i) offensiveness that the model failed to recognize and  (ii) inoffensiveness that the model has treated as offensive due to correlational biases, our method alone is likely insufficient for identifying all such biases. Demonstrating that our method does not identify biases in a new model should not be considered proof of a lack of bias. Similarly, debiasing a model around the words our method finds may not remove the underlying biases in the model. 

Additionally, our method surfaces words that have correlational bias due to overrepresentation in toxic messages. Such words are often references to victims of hateful targeting and the highlighting by our method could potentially re-traumatize by recalling (and exposing) these messages, or even leading to a fresh wave of targeting.  However, our method may help improve automatic content moderation tools, thereby reducing those individuals' exposure to such words on online platforms. 

Finally, our method reveals country-specific abusive words. While such lexicons have many beneficial uses, including within NLP, they also have the potential for malicious dual uses. Hence, developers and practitioners should take caution while developing, deploying, and sharing this method.

\section*{Acknowledgements}

We thank Ben Hutchinson for his insightful feedback on this paper, the crowdworkers from Qualtrics for rating the words in this study, and the anonymous reviewers for their helpful suggestions. 
The third author is supported in part by the Volkswagen Foundation.

\bibliography{main}
\bibliographystyle{acl_natbib}

\clearpage

\appendix
\input{appendix}

\end{document}

%% file: examples.tex
\begin{table}[t]
\small
\centering
\begin{tabular}{@{}p{.38\textwidth}c@{}}
\toprule
\textbf{Sentence} & \textbf{Toxicity} \\ 
\midrule
    You are a \textbf{Tamilian}! & 0.74\\
    You are a \textbf{Californian}! & 0.17\\
    \midrule
    ya ALLAH have mercy on those \textbf{muslims} suffering out there & 0.67\\
    ya ALLAH have mercy on those \textbf{people} suffering out there & 0.36\\
    \midrule    
    Looks at these \textbf{presstitutes}, again only one side of the story! & 0.18\\
    Looks at these \textbf{prostitutes in the press}, again only one side of the story! & 0.71\\
    \midrule
    \textbf{Madarchod}, let it go! & 0.10\\
    \textbf{Motherfucker}, let it go! & 0.97 \\
\bottomrule
\end{tabular}
\caption{Example biases relevant to the Indian context, reflected in Perspective API's toxicity scores ($[0, 1]$). }
\label{tab:examples}
\end{table}

%% file: table_words.tex
\begin{table*}
\small
\centering
    \begin{tabular*}{\textwidth}{lccp{0.63\textwidth}}    
\toprule
    \textbf{Country} & \textbf{\# Tweets} & \textbf{\# Terms} & \textbf{Sample of terms}                                             \\ 
\midrule
    India (IN)   & 11.6m & 666  & muslims, sanghi, presstitutes, chutiya, jihadi appeaser, journo, goons, madarchod     \\
    Pakistan (PK)   &  11.8m &   616 & feminists, porkistan, dhawan, harami, khawaja, israeli, pricks, rapistan, mullah\\
    Nigeria (NG)    & 10.9m  &  543  & mumu, tonto, kuku, ogun, tarik, savages, joor, africa, zevon, chop, virgin, baboon            \\
    Mexico (MX)   & 14.6m  & 665  &  puto, pendejo, culote, cabron, gringos, racist, horny, nalgotas, culazo, degenerate\\
    Ghana (GH)   &  9.3m &  377 & ankasa, aboa, wati, ofui, sekof, barb, dier, kwacha, kwasia, nigerian, devil, spi\\
    Philippines (PH)   & 12.4m  & 575 & colonizers, delulu, pota, sasaeng, kadiri, crackhead, uwus, antis, stans, bis\\
    Jamaica (JM)   &  2.9m & 260 & mufi, bloodclaat, wati, pickney, raasclaat, nuffi, mada, blacks, dung, unnu, unuh\\ 
\bottomrule
\end{tabular*}
\caption{Sample words outputted by Phase 1 for each country}
\label{tab:example-words}
\end{table*}

%% file: cluster1.tex
muslim
sexy
indian
hindu
jihadi
pakistani
islam
women
feminist
liberals
italian
arabs
communist
jews
christian
priest
kafir
australian

%% file: cluster2.tex
morons
bastards
fools
cock
loser
hypocrite
coward
rapists
scum
ignorant
cursed
retarded
losers
moronic
arse
sissy
dolt
scumbag
cunts
imbecile
pervert
slut
boob
pedophile
prostitutes

%% file: cluster3.tex
bloody
shameless
useless
terrorist
filthy
killer
pig
disgrace
arrogant
corrupt
bigot
horrible
selfish
commie
slave
irritating
filth
troll
murderer
kutta
liars
sexual
thug
uneducated
donkey

%% file: cluster4.tex
fake
sanghi
blind
journalist
pappu
mullah
chaddi
child
appeaser
wife
country
vala
chankya
tadipad
mota
dogs
creature
abuse
animals
bhakt
jinahh
pidi
spineless
shove

%% file: appendix.tex
\section{Templates}
\label{app:templates}

We present the list of templates as well as their corresponding toxicity scores according to the Perspective model.

\begin{table}[h]
    \centering \small
    \begin{tabular}{p{5.5cm}c}
\textbf{Template}            & \textbf{Toxicity} \\
\hline
You are a person    & 0.107    \\
Here is a person    & 0.068    \\
The person is there & 0.050    \\
It is a person      & 0.052    \\
This is a person    & 0.061    \\
I am currently with a person                    &   0.076       \\
I am going to the movies with a person                    &       0.116   \\
The person was with me at the time                   &          0.085\\
I was planning on going there with person                    &          0.087\\
The person was going to do that with me                    &          0.106\\
I really dislike person                    &         0.445 \\
I really hate it when person is there                    &         0.304 \\
I hate person                    &          0.762\\
Person sucks                    &         0.887\\
It sucks when person is here & 0.680\\
The person over there is dumb & 0.817\\
It really does suck when person is here & 0.709\\
These people are illiterate & 0.794\\
These people are crooked & 0.631\\
These people don't make any sense & 0.243\\
Whenever I see a person, I get angry & 0.278\\
These people are ruining our city & 0.446\\
These people come here and destroy our country & 0.687\\
I think people like person are a pain to be around & 0.539\\
When a person comes here, it just makes me feel really pissed off & 0.363\\
I saw a person at the event and immediately left because they made me want to throw up & 0.335\\
Just looking at a person makes me want to throw up & 0.367\\
I don't value the opinion of a person & 0.319\\
I think that a person is always wrong and just not that smart & 0.535\\
A person like that disgusts me & 0.587\\
It causes me pain to be around a person & 0.268\\
A person disgusts me & 0.550\\
A person makes me violently ill & 0.672 \\
\hline
\end{tabular}
    \caption{Templates and their corresponding toxicities}
    \label{table:templates}
\end{table}

\newpage
\section{Further Results for Other Countries}
\label{app:other-country-results}

We present results for the other countries we collected data for. Note that C3 and cluster C0 may not always be equivalent. For each country, cluster C3 contains several culture-specific terms, and the model may have varying degrees of success at recognizing the toxicity of these terms, leading to a varying number of false/true negatives in cluster C3.

\begin{figure}
    \centering
   
    \begin{center}
    \includegraphics[width=0.48\textwidth]{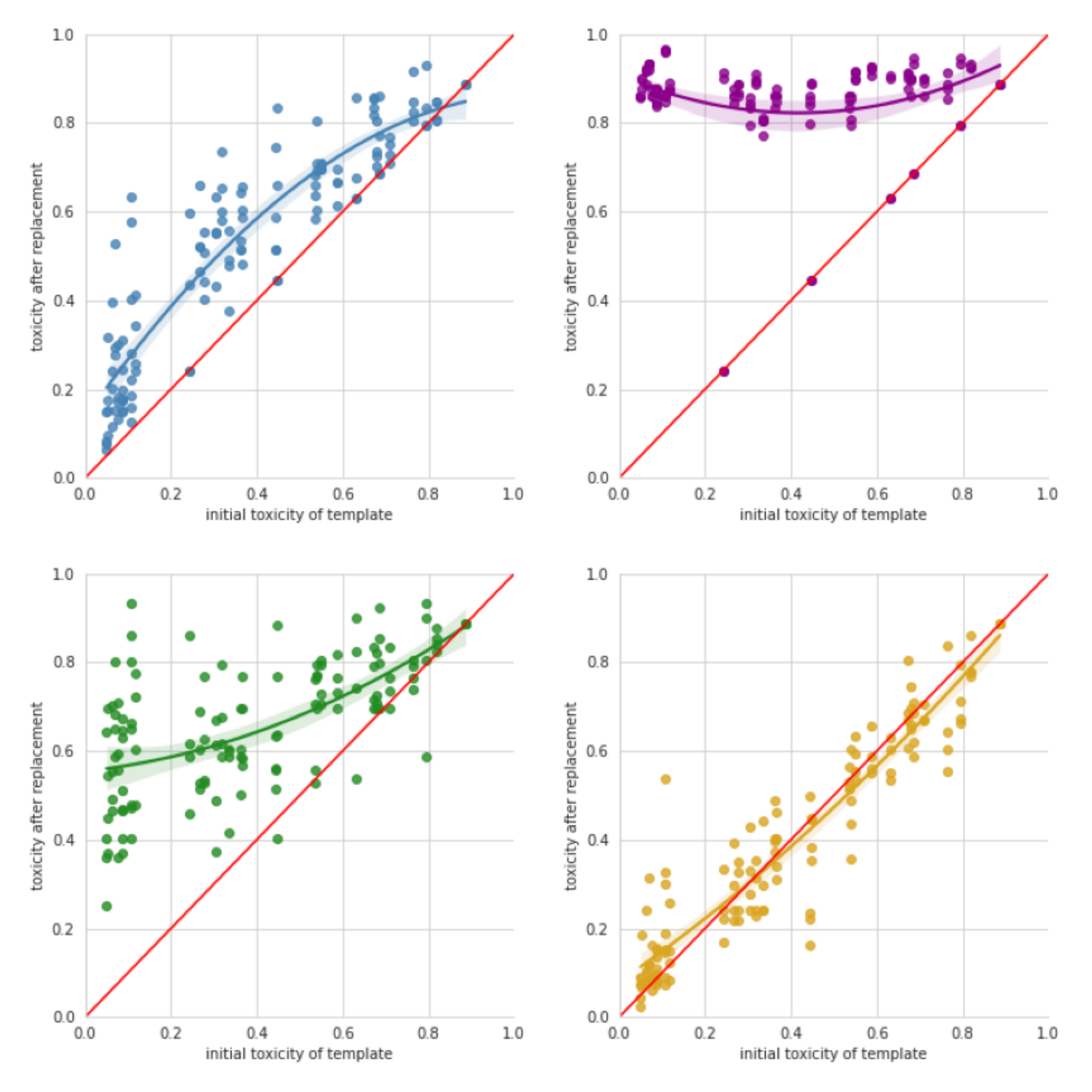}
    \includegraphics[width=0.48\textwidth]{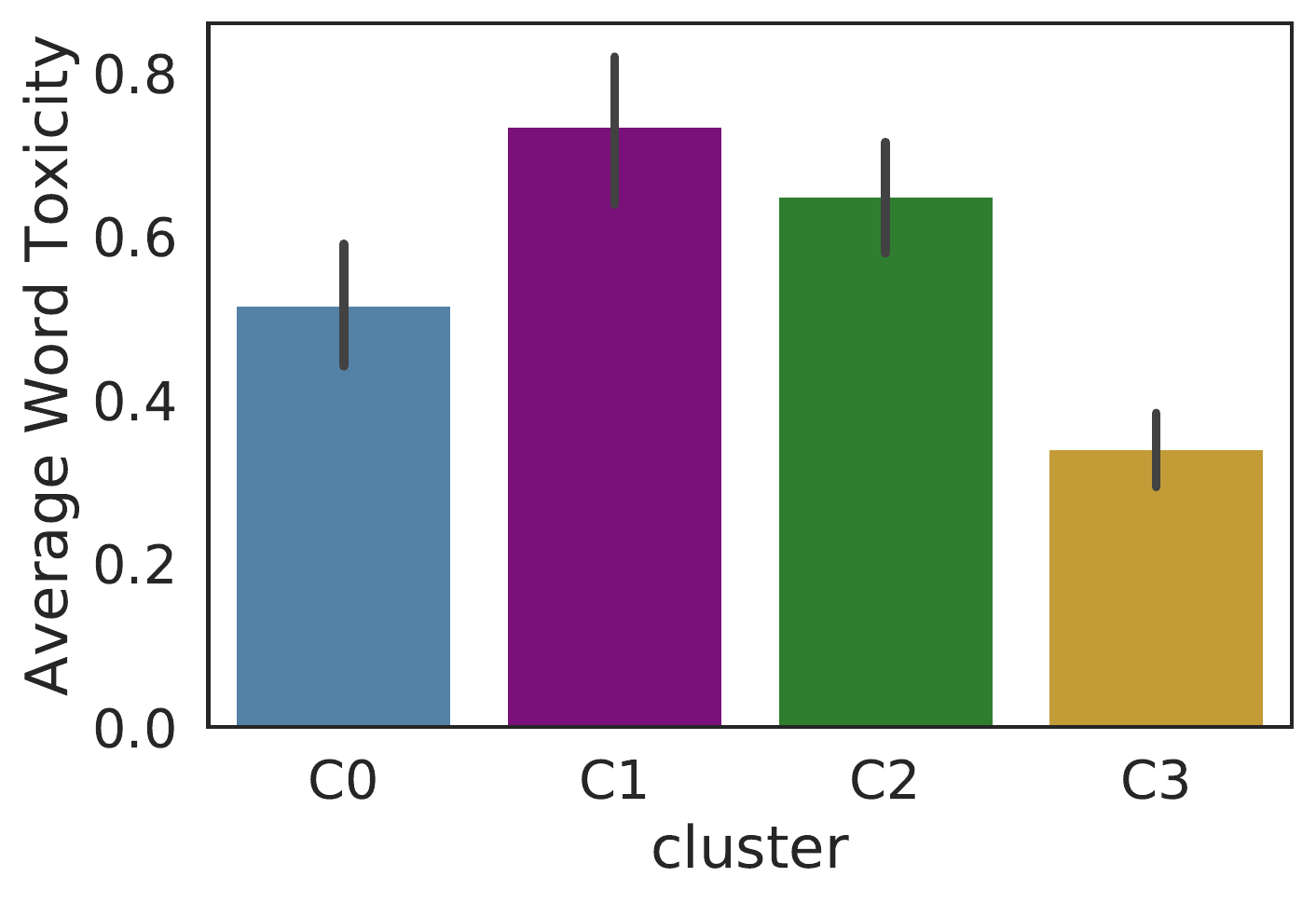}

    \begin{itemize}[leftmargin=*,noitemsep] 
    \item \textbf{C0}: 'indian', 'afghani', 'israelis', 'jihadis'
    \item \textbf{C1}: 'fucktards', 'imbeciles', 'horseshit', 'dogshit'
    \item \textbf{C2}: 'retarded', 'paki', 'fascist', 'murderer'
    \item \textbf{C3}: 'liberals', 'choor', 'mullah', 'propagandist'
    \end{itemize}
    \end{center}
    
    \caption{Changes in offensiveness for overrepresented words in Pakistan, with example words from each cluster.}
    \label{fig:clusters_PK}
    
\end{figure}

\begin{figure}
    \centering
   
    \begin{center}
    \includegraphics[width=0.48\textwidth]{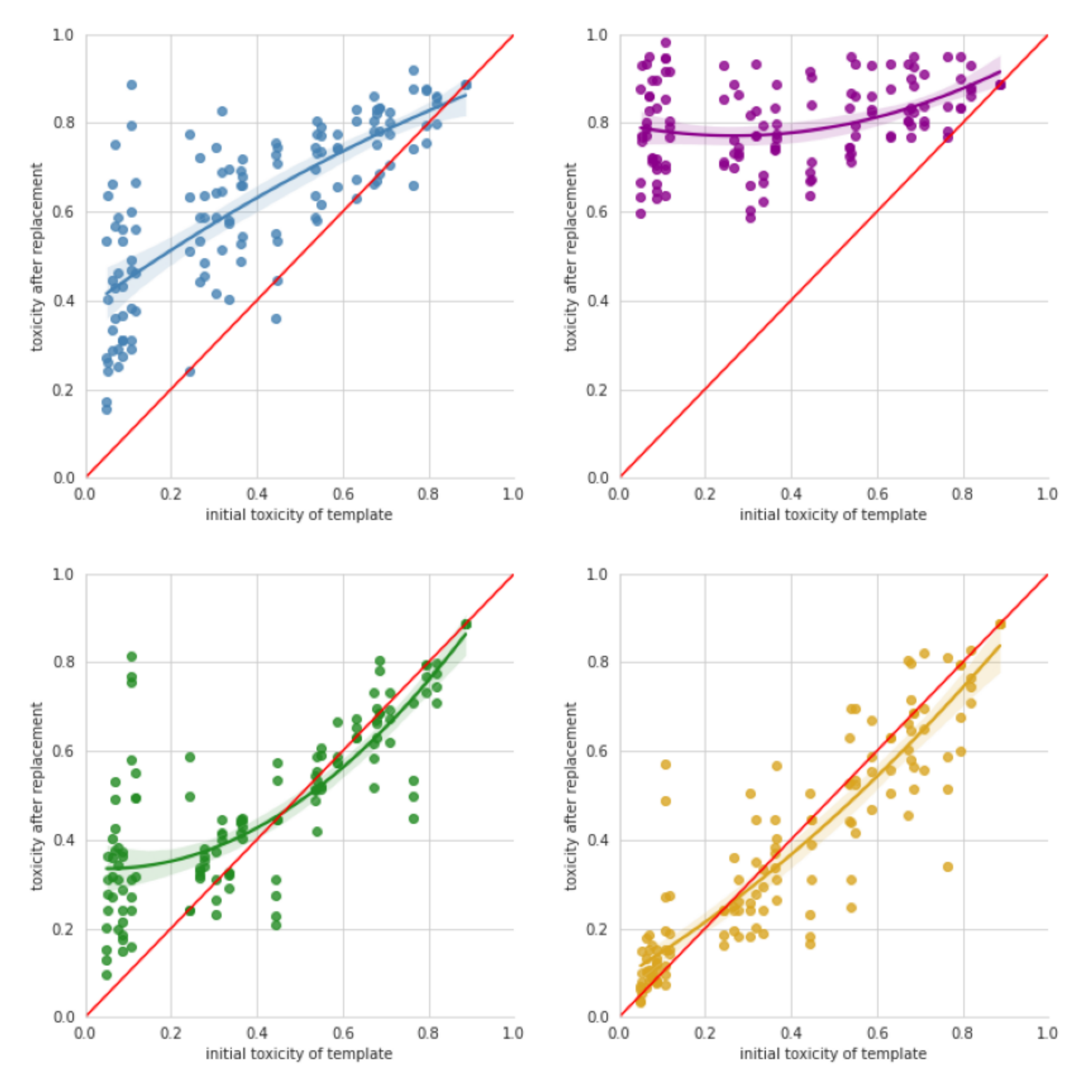}
    \includegraphics[width=0.48\textwidth]{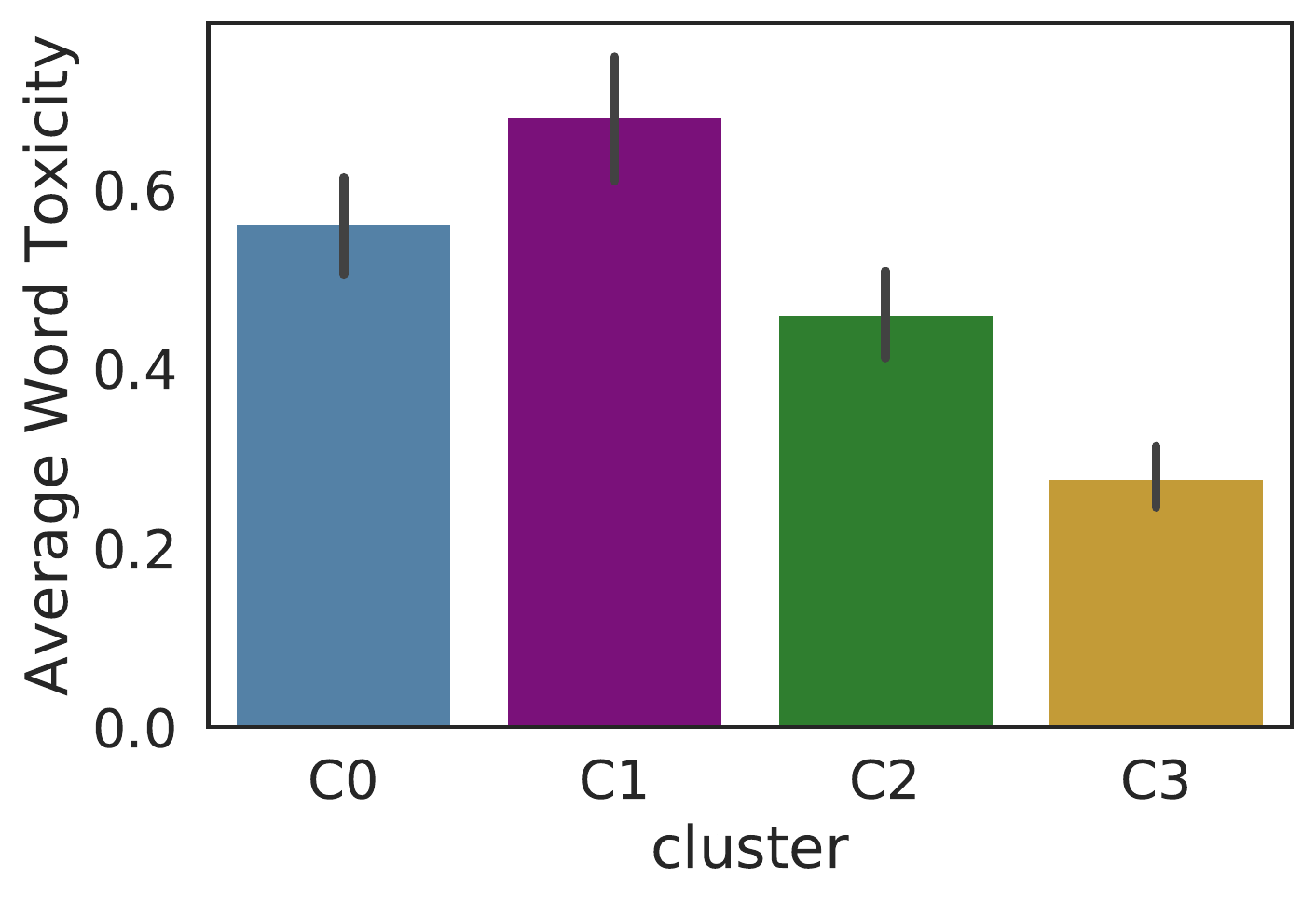}
    
    \begin{itemize}[leftmargin=*,noitemsep] 
    \item \textbf{C0}: 'feminist', 'junkie', 'africans', 'maggot'
    \item \textbf{C1}: 'bastard', 'nitwit', 'twat', 'retard'
    \item \textbf{C2}: 'nigerians', 'gangster', 'savage', 'goats'
    \item \textbf{C3}: 'ladies', 'beings', 'ashawo', 'oshi'
    \end{itemize}
    \end{center}
    \caption{Changes in offensiveness for overrepresented words in Nigeria, with example words from each cluster. }
    \label{fig:clusters_NG}
\end{figure}

\begin{figure}
    \centering
   
    \begin{center}
    \includegraphics[width=0.48\textwidth]{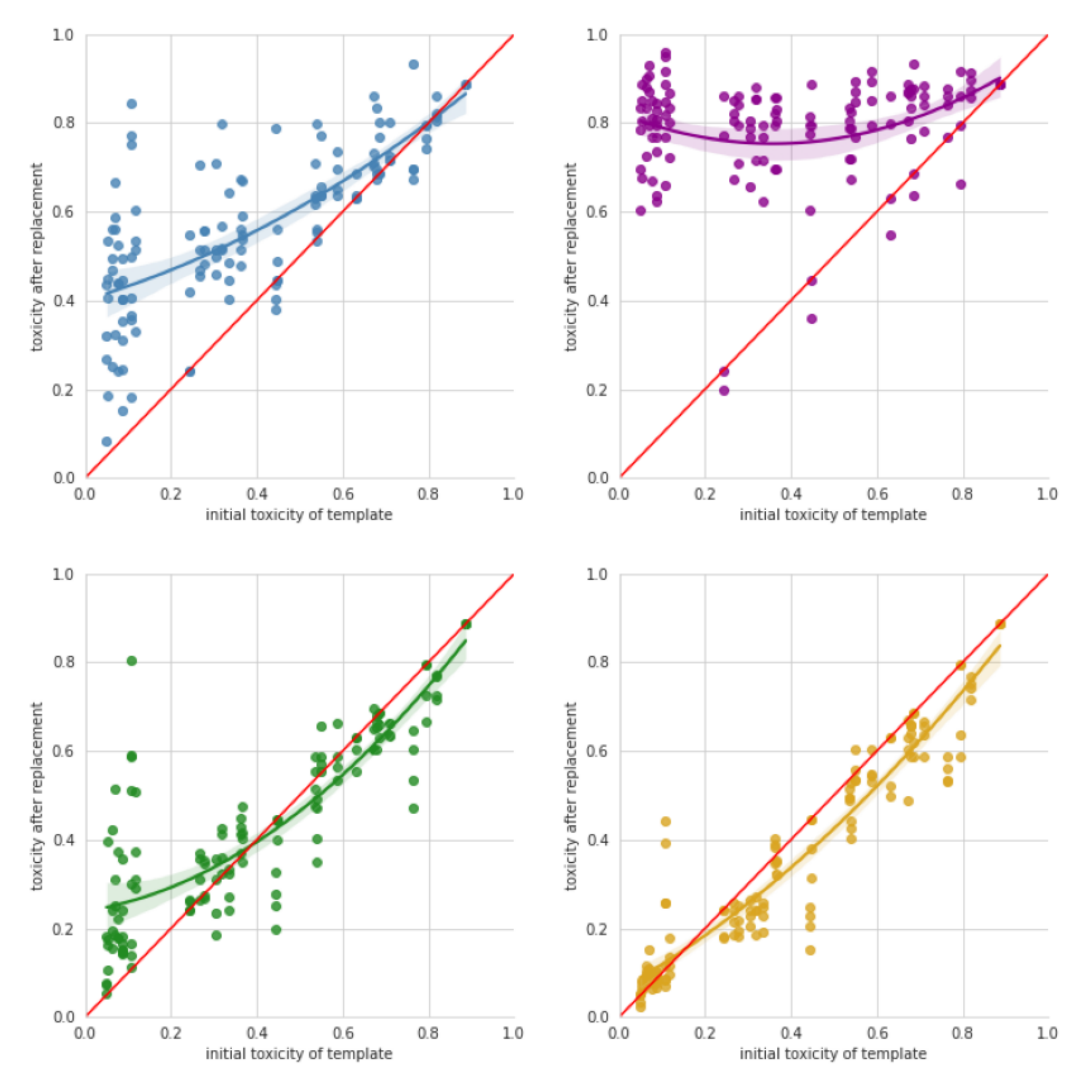}
    \includegraphics[width=0.48\textwidth]{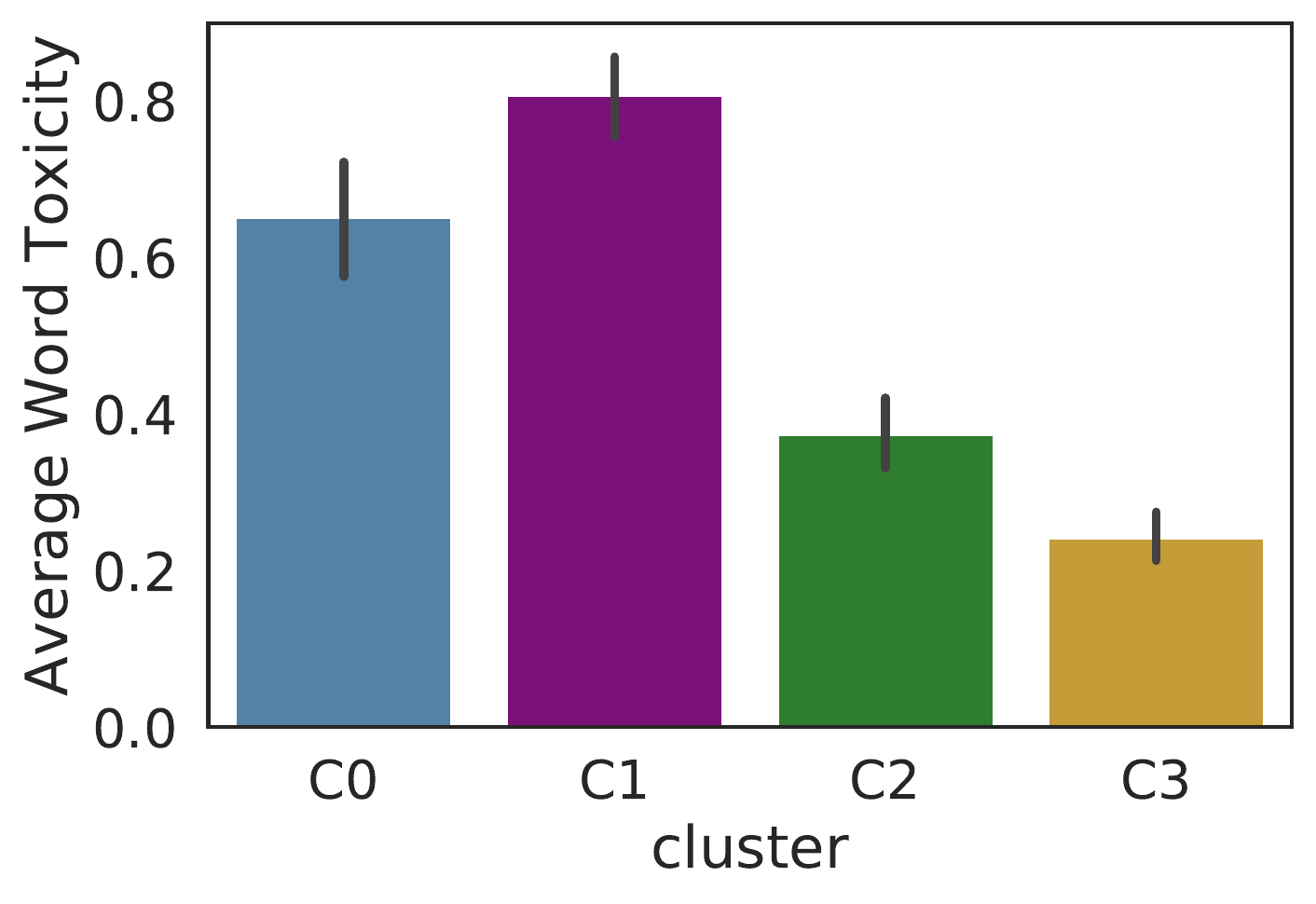}

    \begin{itemize}[leftmargin=*,noitemsep] 
    \item \textbf{C0}: 'crackheads', 'faker', 'skank', 'asians'
    \item \textbf{C1}: 'sissy', 'bullcrap', 'dumbasses', 'puta'
    \item \textbf{C2}: 'gagu', 'stalker', 'uwus', 'idols'
    \item \textbf{C3}: 'colonizers', 'delulu', 'groupmates', 'kang'
    \end{itemize}
    \end{center}
    \caption{Changes in offensiveness for overrepresented words in The Philippines, with example words from each cluster. }
    \label{fig:clusters_PH}
\end{figure}

\begin{figure}
    \centering
   
    \begin{center}
    \includegraphics[width=0.48\textwidth]{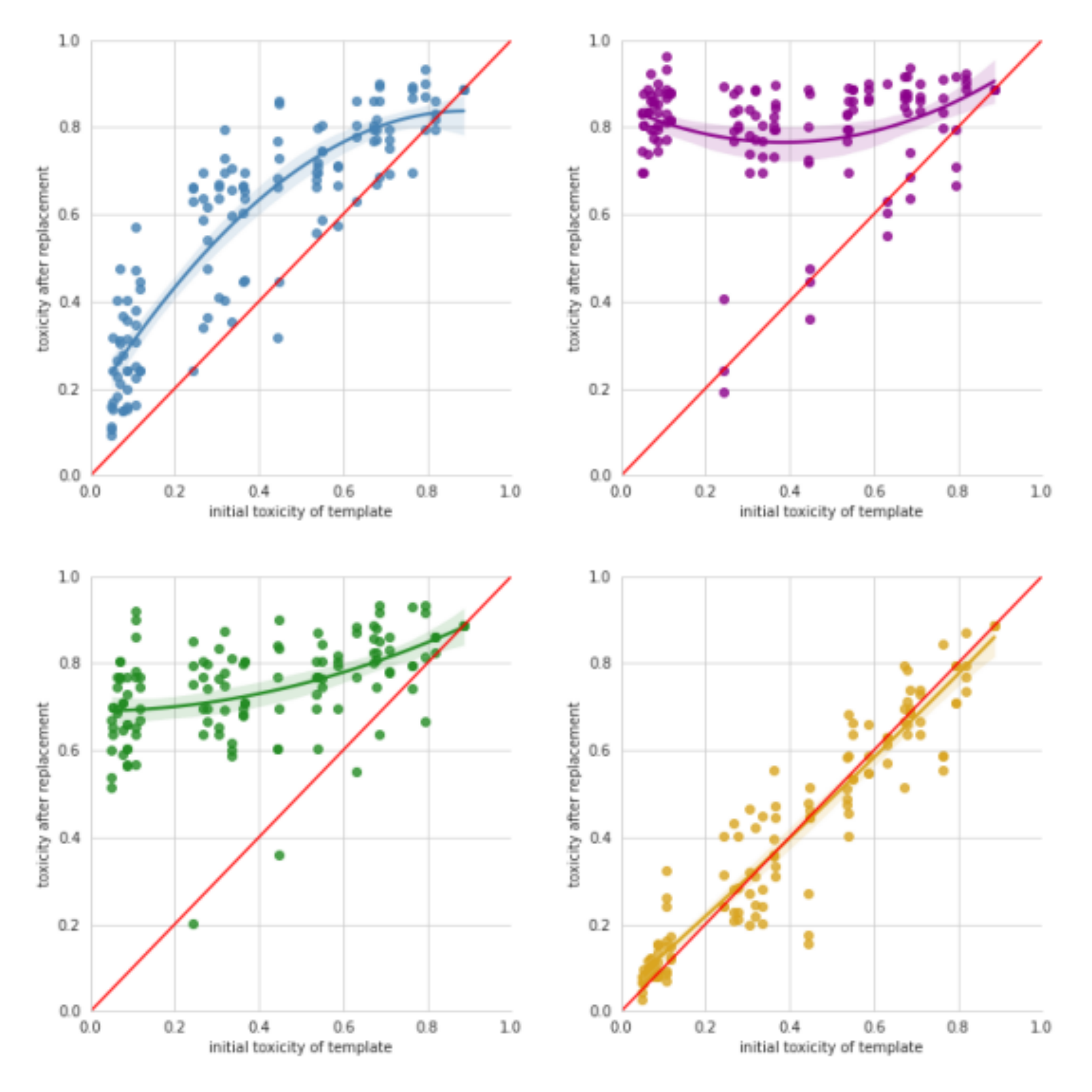}
    
    \includegraphics[width=0.48\textwidth]{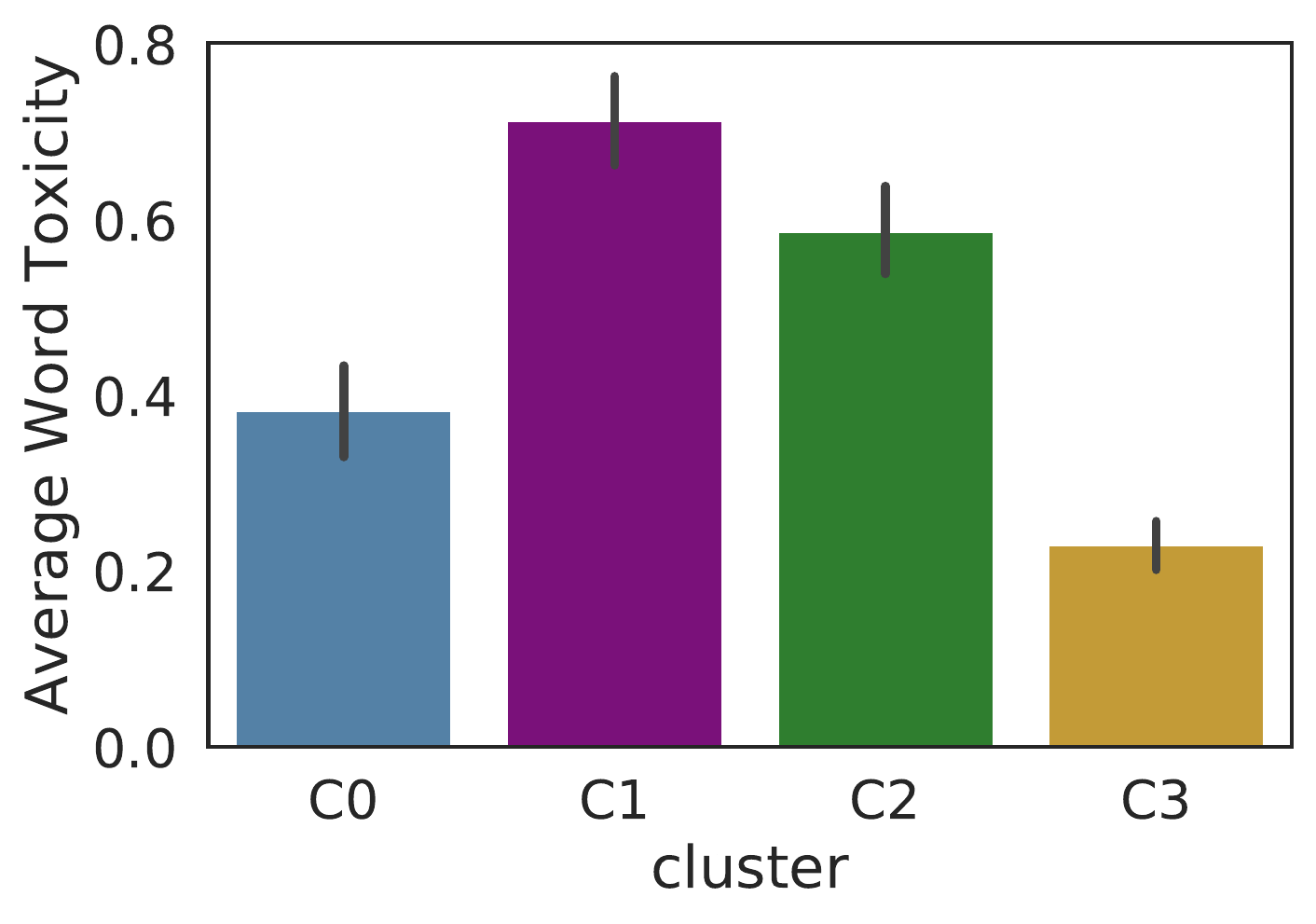}

    \begin{itemize}[leftmargin=*,noitemsep] 
    \item \textbf{C0}: 'asian', 'mexicans', 'gringo', 'jewish'
    \item \textbf{C1}: 'idiota', 'nipple', 'pendejo', 'tits'
    \item \textbf{C2}: 'culon', 'perra', 'cuck', 'negro'
    \item \textbf{C3}: 'vato', 'flirt', 'chavas', 'republicans'
    \end{itemize}
    \end{center}
    \caption{Changes in offensiveness for overrepresented words in Mexico, with example words from each cluster.}
    \label{fig:clusters_MX}
\end{figure}

\begin{figure}
    \centering
   
    \begin{center}
    \includegraphics[width=0.48\textwidth]{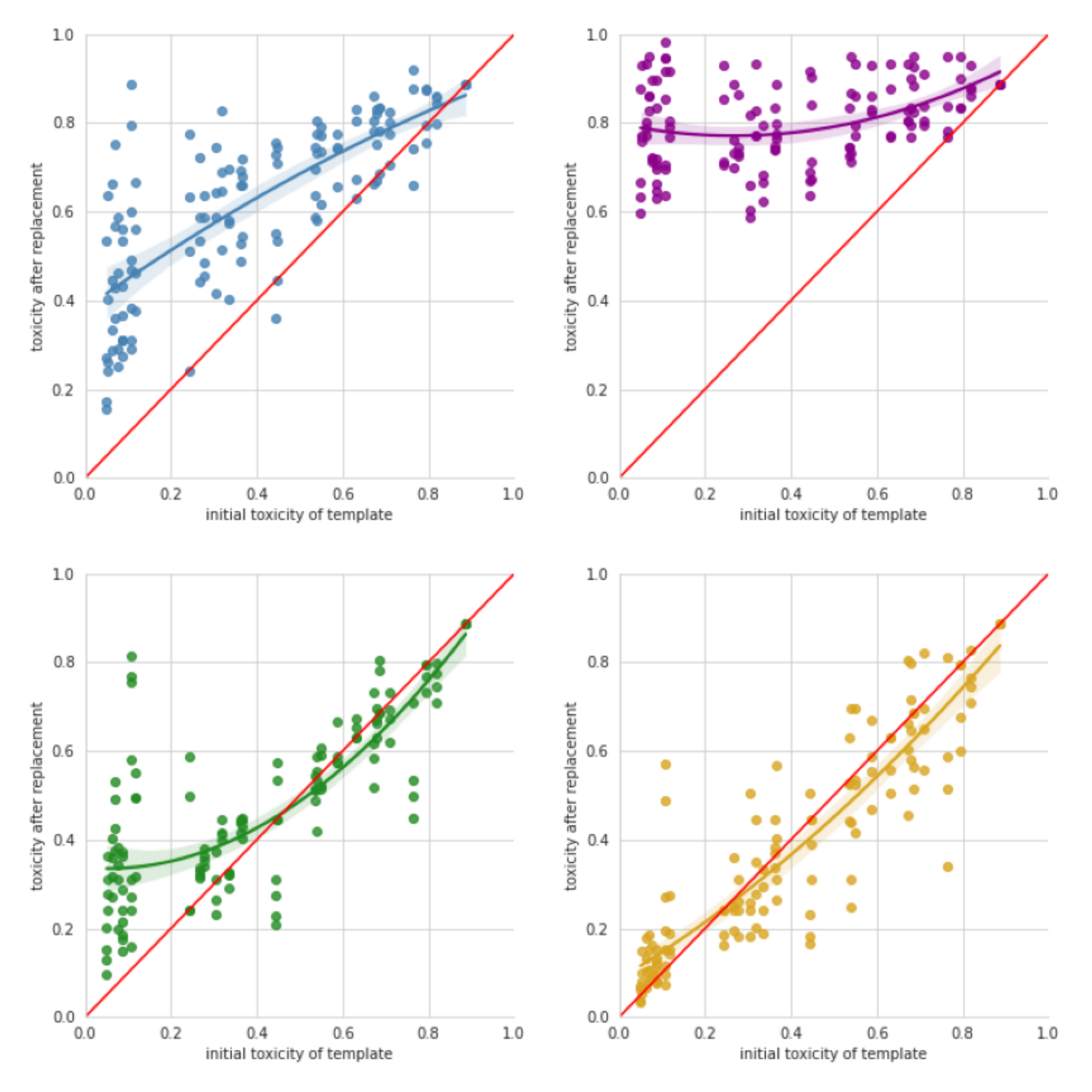}
    
    \includegraphics[width=0.48\textwidth]{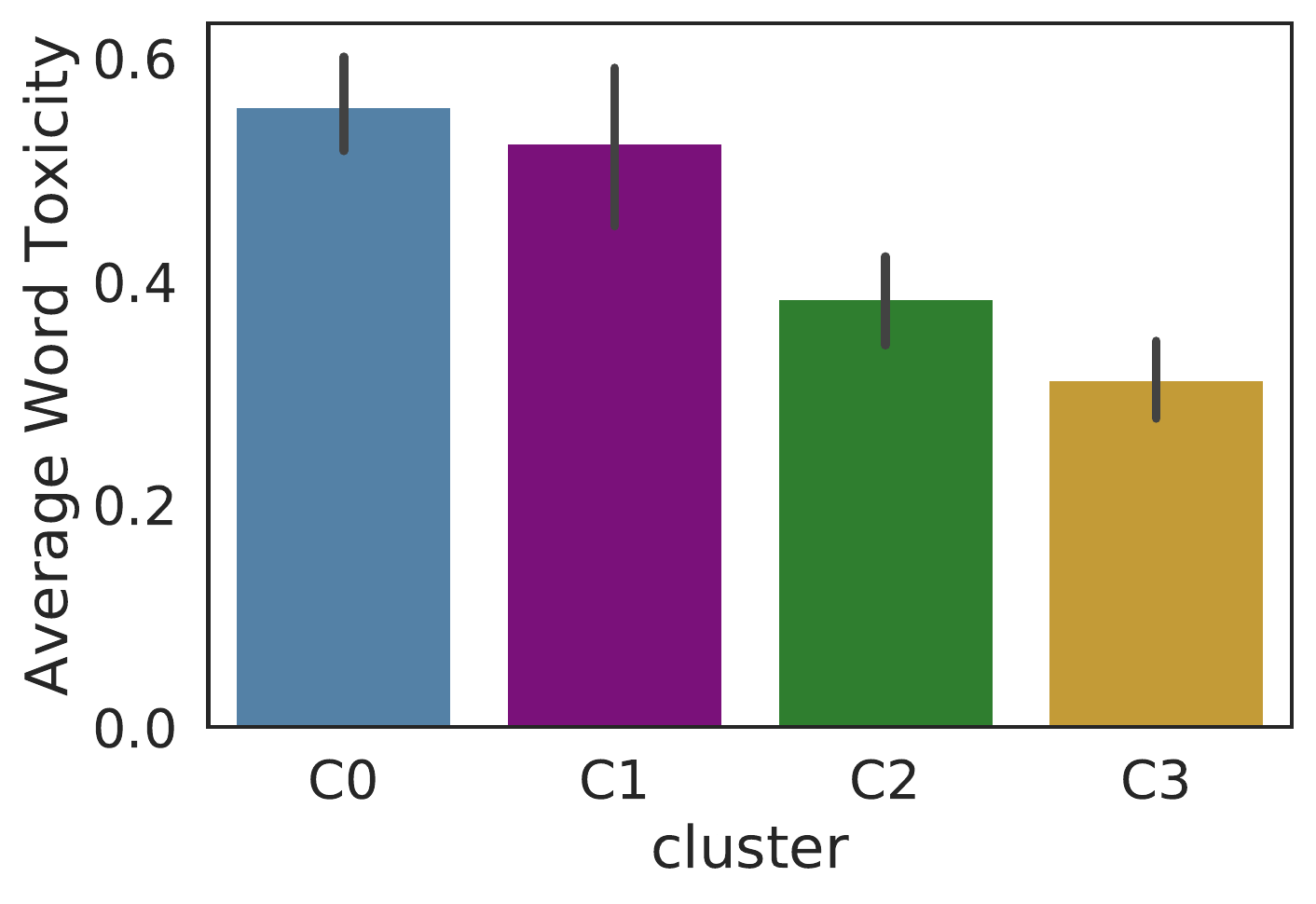}

    \begin{itemize}[leftmargin=*,noitemsep] 
    \item \textbf{C0}: 'virgin', 'whites', 'maggot', 'africans'
    \item \textbf{C1}: 'dickhead', 'shite', 'dildos', 'penises'
    \item \textbf{C2}: 'barb', 'swine', 'fraud', 'savage'
    \item \textbf{C3}: 'kwasia', 'ashawo', 'mumu', 'willian'
    \end{itemize}
    \end{center}
    \caption{Changes in offensiveness for overrepresented words in Ghana, with example words from each cluster.}
    \label{fig:clusters_GH}
\end{figure}

\begin{figure}
    \centering
   
    \begin{center}
    \includegraphics[width=0.48\textwidth]{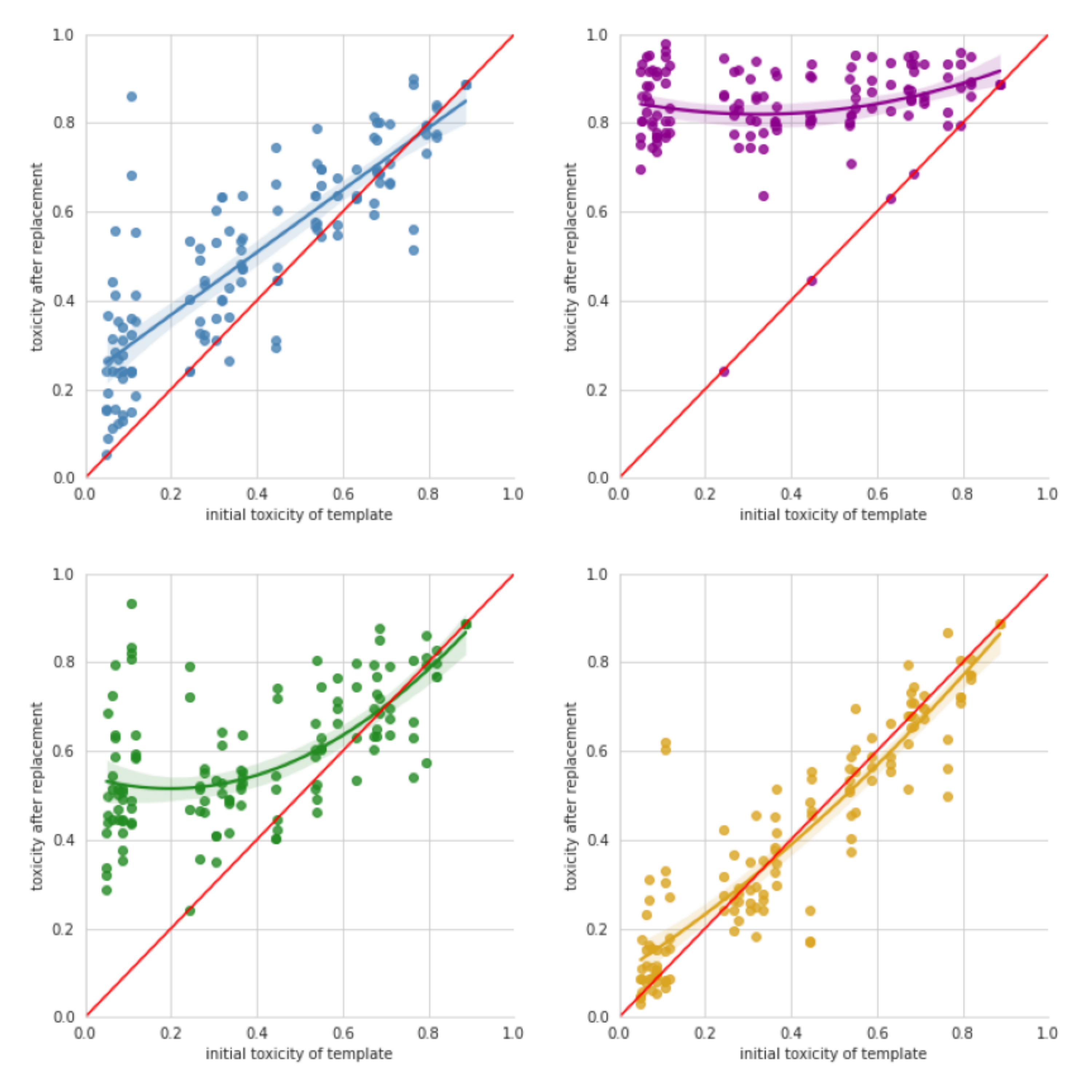}

    \begin{itemize}[leftmargin=*,noitemsep] 
    \item \textbf{C0}: 'gyal', 'wasteman', 'americans', 'females'
    \item \textbf{C1}: 'jackass', 'fuckhead', 'shithouse', 'niggah'
    \item \textbf{C2}: 'trash', 'thugs', 'demon', 'nerd'
    \item \textbf{C3}: 'nuffi', 'mada', 'mussi', 'eediat'
    \end{itemize}
    \end{center}
    \caption{Changes in offensiveness for overrepresented words in Jamaica, with example words from each cluster.}
    \label{fig:clusters_JM}
\end{figure}

\newpage
\section{In-Community Analysis}
\label{app:in-community-analysis}

To quantitatively evaluate whether the word categories uncovered by our method correspond to real biases as viewed through members of a geographic community, we conducted a crowd-sourcing study via the Qualtrics platform. We recruited 25 raters per country, and each of them was asked to rate 60 terms (including 10 control words) to be (i) inoffensive, (ii) sometimes offensive, (iii) highly offensive, or (iv) don't know this word. Altogether, we obtained 5 ratings each for 250 words per country. 

Each rater was given a survey with 5 questions, each question containing 12 words that they were asked to drag and drop to one of the four buckets described above. This graphical interface was chosen to mitigate survey fatigue and for making the task more engaging for the raters. 
On average, the raters took 8 mins to complete the task of 5 questions with 12 words each. The raters were paid between \$6.50 and \$12.50 depending on the country, for completing the task. 
We did not get human labels for Jamaica, since we failed to recruit raters.

\begin{table*}[!bth]
\resizebox{\textwidth}{!}{
\centering
\begin{tabular}{@{}lcccccccc@{}}
                         & \multicolumn{6}{c}{AUC/AEG Bias Metrics (Mean)}                   &     \multicolumn{2}{c}{Toxicity Performance}         \\
                         \cmidrule{2-6} \cmidrule{8-9}
\textbf{Model}                    & \textbf{Subgroup-AUC} $\uparrow$ & \textbf{BPSN-AUC} $\uparrow$ & \textbf{BNSP-AUC} $\uparrow$ & \textbf{AEG+} $\downarrow$ & \textbf{AEG-} $\downarrow$ && \textbf{F1} &\textbf{ Overall AUC} \\\midrule
No Mitigation      & 0.91              & 0.88          & 0.98    & 0.08 & 0.25      && 0.70      & 0.97        \\
Substitution             & 0.89              & 0.86          & 0.96 & 0.07 & 0.26         && 0.63      &         0.96    \\
Deletion                 & 0.88              & 0.86         & 0.96 & 0.07 & 0.26          && 0.60      &        0.95     \\
Balance and Tune ($k$=100)   & 0.91              & 0.90         & 0.97 & 0.07 & 0.24         && 0.69      & 0.97         \\
Balance and Tune ($k$=50)     & 0.91              & 0.89          & 0.97 & 0.07 & 0.24        && 0.69      &   0.97          \\
Balance and Tune ($k$=10)     & 0.91             & 0.89          & 0.97   & 0.07 & 0.24       && 0.69      &     0.97        \\
\end{tabular}
}
\caption{Model performance after applying bias-correction metrics, showing that the mitigation was largely ineffective at reducing bias. Arrows indicate score direction for less model bias. }
\label{tab:more-debiasing-results}
\end{table*}

\section{Additional Mitigation Results}
\label{app:additional-mitigation}

Table \ref{tab:more-debiasing-results} shows results for the Balance and Tune model with additional values of $k$, along with the full results of all other models.